\documentclass[10pt,twocolumn,letterpaper]{article}

\usepackage[pagenumbers]{cvpr}
\usepackage{booktabs}  
\usepackage{arydshln}  
\usepackage[inkscapelatex=false]{svg}
\usepackage[accsupp]{axessibility}  

\usepackage{caption}
\usepackage{subcaption}
\usepackage[pagebackref,breaklinks,colorlinks,allcolors=cvprblue]{hyperref}
\usepackage[capitalize]{cleveref}
\crefname{section}{\S}{\S}
\Crefname{section}{Section}{Sections}
\Crefname{table}{Table}{Tables}

\usepackage[ruled,vlined]{algorithm2e}
\usepackage{colortbl}
\definecolor{hlcell}{HTML}{D6EAF8}   

\newcommand{\TODO}[1]{\textbf{\color{red}[TODO: #1]}}
\newcommand{\nknote}[1]{\textbf{\color{green}[nilesh: #1]}}
\newcommand{\dima}[1]{\textbf{\color{orange}[dima: #1]}}

\renewcommand{\TODO}[1]{}

\renewcommand{\nknote}[1]{}
\renewcommand{\dima}[1]{}

\definecolor{cvprblue}{rgb}{0.21,0.49,0.74}

\def\paperID{*****}
\def\confName{CVPR}
\def\confYear{2026}

\title{Less is More: Data-Efficient Adaptation for Controllable Text-to-Video Generation}

\author{
Shihan Cheng$^{1}$ \quad Nilesh Kulkarni$^{2}$ \quad David Hyde$^{1}$ \quad Dmitriy Smirnov$^{2}$ \\
$^{1}$Vanderbilt University, USA \quad
$^{2}$Netflix, USA \\
{\tt\small \{shihan.cheng, david.hyde.1\}@vanderbilt.edu} \quad
{\tt\small \{nkulkarni, dimas\}@netflix.com}
}
\begin{document}

\twocolumn[{%
\renewcommand\twocolumn[1][]{#1}%
\maketitle
{\centering
\includegraphics[width=0.95\textwidth]{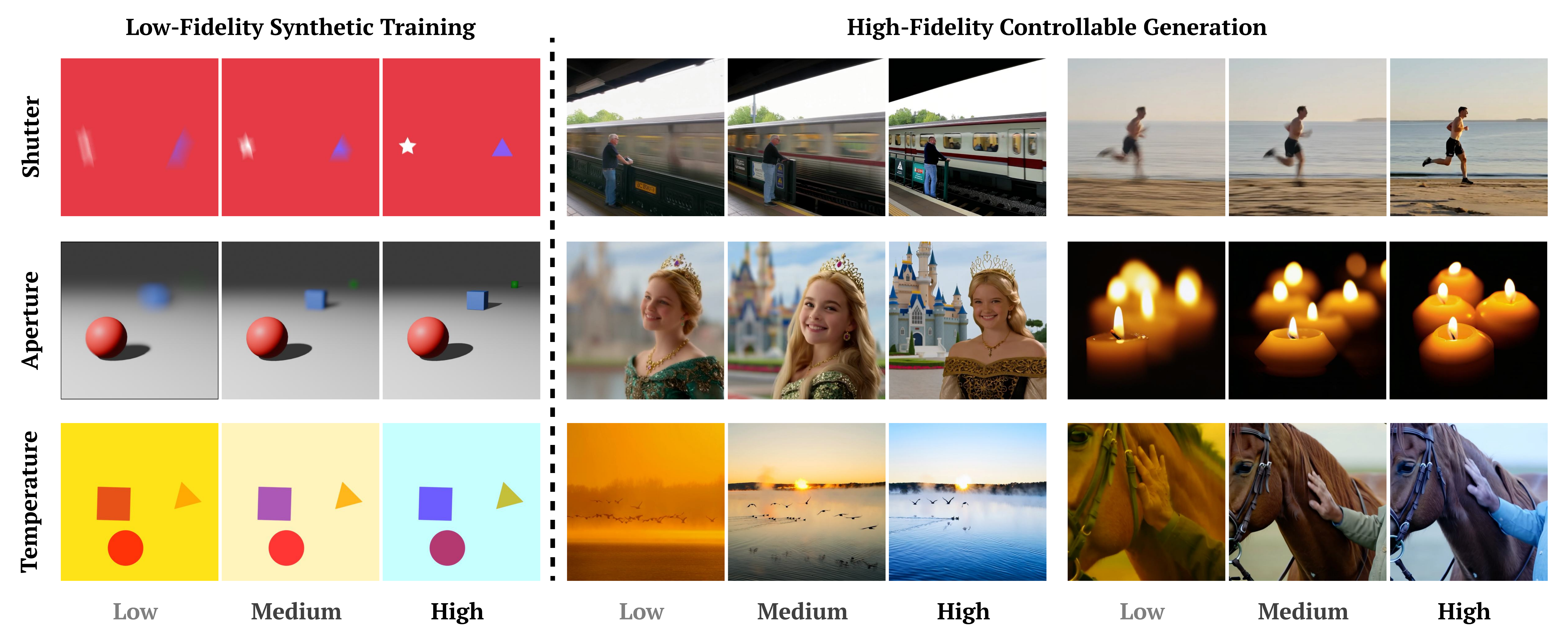}
}
\captionof{figure}{
\textbf{Our ``Less is More'' framework for data-efficient controllable generation.}
A T2V backbone, fine-tuned solely on a sparse, low-fidelity synthetic dataset (left), learns to generalize to complex physical controls.
This enables precise, high-fidelity manipulation of shutter speed (motion blur), aperture (bokeh), and color temperature during real-world inference (right), driven by a continuous control.
\vspace{1em}}
\label{fig:teaser}
}]
\begin{abstract}
Fine-tuning large-scale text-to-video diffusion models to add new generative controls, such as those over physical camera parameters (e.g., shutter speed or aperture), typically requires vast, high-fidelity datasets that are difficult to acquire.
In this work, we propose a data-efficient fine-tuning strategy that learns these controls from sparse, low-quality synthetic data. We show that not only does fine-tuning on such simple data enable the desired controls, it actually yields superior results to models fine-tuned on photorealistic ``real" data. Beyond demonstrating these results, we provide a framework that justifies this phenomenon both intuitively and quantitatively.
\end{abstract}
\section{Introduction}
Recent advances in generative AI using diffusion models have enabled unprecedented levels of quality in video generation.
The primary foundational models for video generation are text-to-video (T2V) such as~\cite{wan2025wan, polyak2024movie}, where users describe their desired creations in natural language. Due to limitations of controllability via text, significant effort has been put into creating methods that accept other input modalities, such as images, keyframes, depth maps, bounding boxes, pose skeletons, driver videos, camera trajectories, and so on~\cite{wan2025wan, jiang2025vace, cheng2025wan, bai2025recammaster}.
However, achieving consistent, reliable, and intuitive fine-grained control over all aspects of the output video remains a challenge.
In this work, we tackle the problem of adding a control mechanism over specific low-dimensional physical or optical properties, such as camera intrinsics, via a simple synthetic data generation and model fine-tuning framework.\footnote{Code and models are publicly available at \href{https://github.com/csh-apprentice/Less_Is_More}{\url{https://github.com/csh-apprentice/Less_Is_More}}.}

A common pattern that has emerged in methods for producing specialized generative video models is starting with a large ``foundation'' model, trained on huge amounts of video data, and fine-tuning it on a carefully crafted task-specific smaller dataset~\cite{polyak2024movie}.
Such datasets help the model focus on a particular character identity, artistic style, or specialized effect.
The success of such approaches hints at the fact that the initial pre-training equips the model with many useful priors implicit in its latent representation, which can be explicitly ``coaxed'' out during post-training. Ours is the first approach that aims to enable conditioning on camera effects (like shutter speed, focal length, etc) in pre-trained video generative models for consistent generation. While the quality of the post-training data is indeed critical, we argue and demonstrate that data which are perfectly photorealistic and representative of the output domain can, surprisingly, be not only unnecessary but even detrimental for certain specializations.

To this end, we propose a framework that leverages synthetic data to learn multiple conditioning effects.
We contribute a novel joint training approach that factorizes adaptation: a standard low-rank adapter (LoRA)~\cite{hu2022lora} encodes a minimal domain shift, while a disentangled cross-attention adapter learns the conditioning physical effect.
We additionally contribute a formal analysis of \emph{why} this data-efficient approach succeeds.
We demonstrate that, counterintuitively, the success of small-data fine-tuning hinges on the \emph{simplicity} of the synthetic data.
We show that fine-tuning on photorealistic synthetic data, while seemingly a higher-fidelity choice, induces catastrophic forgetting by corrupting the backbone's pre-trained priors, leading to a ``content collapse.''
To quantify this, we introduce a new evaluation framework that measures this generative drift and its impact on semantic fidelity.
Using this framework, we show that a model trained on simple data retains its generative diversity and high semantic fidelity, while the model trained on complex data suffers a quantifiable and catastrophic collapse.
Our work provides both a data-efficient method for controllable video generation and a formal methodology for diagnosing and preventing backbone corruption during adaptation.

\section{Related Work}

\noindent\textbf{Text-to-Video Generation.}
Diffusion models~\cite{ho2020denoising,sohl2015deep} have become the leading approach for video generation, largely by adapting pre-trained text-to-image models with temporal modules. Early methods like AnimateDiff~\cite{guo2023animatediff} showed that adding temporal attention to a frozen T2I U-Net~\cite{ronneberger2015u} enables high-quality motion. The field has since moved toward Diffusion Transformers (DiTs)~\cite{peebles2023scalable}, which underpin state-of-the-art open-source systems such as Hunyuan Video~\cite{kong2024hunyuanvideo} and Wan~\cite{wan2025wan}. Closed-source models including Sora~\cite{OpenAI_Sora_2024}, Gen-3~\cite{Runway_Gen3_2024}, and Kling~\cite{Kuaishou_Kling_2024} further push the frontier with long-form, high-fidelity, physically consistent video. Although our experiments use the Wan2.1 T2V model with three scalar controls, our fine-tuning paradigm and evaluation framework are architecturally compatible with standard DiT-based video diffusion backbones, as they rely solely on standard LoRA layers and a decoupled cross-attention conditioning path; empirical validation on other backbones remains future work.

\vspace{1mm}
\noindent\textbf{Controllable Generative Models.}
Recent work introduces explicit control in generative models through spatial conditions such as depth, masks, sketches, boxes, and segmentation maps~\cite{bhat2024loosecontrol, wang2023imagen, voynov2023sketch, li2023gligen, couairon2022diffedit}. ControlNet~\cite{zhang2023adding} and T2I-Adapters~\cite{mou2024t2i} support such inputs efficiently, with extensions to video focusing on camera pose and trajectory~\cite{he2024cameractrl, yu2024viewcrafter, wang2024motionctrl}. In the image domain, Bokeh Diffusion~\cite{fortes2025bokeh} and Generative Photography~\cite{Yuan_2024_GenPhoto} study camera-aware image synthesis, while CamEdit~\cite{qin2025camedit} targets camera-parameter-controlled photorealistic image editing. However, these image-based settings do not directly address text-to-video adaptation, where temporal consistency and controllable motion must also be preserved. Token-based conditioning via CSaT~\cite{fang2024camera} offers implicit control but remains data-heavy. Large-scale, richly annotated video datasets for physical camera parameters are challenging to obtain. LoRA-based adapters also provide lightweight specialization for related settings, including controllable spatial-temporal video generation~\cite{zhang2025lion}, bokeh control~\cite{fortes2025bokeh}, image restoration~\cite{park2024contribution}, and super-resolution~\cite{sun2025pixel}. Our approach builds on these ideas by injecting scalar physical parameters through a decoupled cross-attention adapter for controllable video synthesis, enabled by targeted synthetic data.

\vspace{1mm}
\noindent\textbf{Learning from Synthetic and Sparse Data.}
The difficulty of large-scale video collection has led to work showing that strong representations can emerge from simple synthetic data~\cite{tian2024learning, mayer2018makes}. Procedurally generated videos can rival natural-video pretraining~\cite{yu2024learning}, and synthetic datasets often serve as effective substitutes for real videos~\cite{guo2022learning, kim2022transferable}. Synthetic 3D environments also provide precise labels for tasks like optical flow, tracking, and video depth~\cite{dosovitskiy2015flownet, zheng2023pointodyssey, yang2024depth}, and recent methods use such data for low-shot motion learning~\cite{ren2024customize, wu2024lamp}. These results suggest that motion priors need not rely on complex natural data, motivating our use of sparse, task-specific synthetic supervision to condition T2V models on camera parameters such as shutter speed, aperture, and temperature.

\noindent\textbf{Evaluation and Metrics for Generative Video.}
Existing one-shot fine-tuning methods \cite{wu2023tune} and even zero-shot approaches \cite{geyer2023tokenflow,qi2023fatezero,yang2023rerender} still struggle to generate high-quality, varied videos.
Even just quantitatively evaluating and comparing against such models remains a complex, multi-dimensional problem.
Common metrics include the Fr\'echet Video Distance (FVD) \cite{unterthiner2018towards,unterthiner2019fvd}, which measures distributional similarity in a learned feature space for action recognition, and CLIP Score \cite{radford2021learning}, which quantifies text–video semantic alignment.
More recent works have proposed adaptations for video, such as X-CLIP Score \cite{ni2022expanding}, or more robust alternatives like VQA Score \cite{lin2024evaluating}, which is less prone to error on complex compositional prompts.
\citet{shuttleworth2025loravsfinetuningillusion} introduce spectral diagnostic tools for analyzing how LoRA and full fine-tuning reshape pre-trained diffusion weights.
While benchmarks such as VBench~\cite{huang2024vbench}---which we adopt in our evaluation---offer a suite of metrics for assessing temporal consistency, object fidelity, and aesthetic quality,
we identify a critical methodological gap: no existing metric quantifies the intrinsic complexity of the fine-tuning data or its impact on backbone drift.
We address this with a novel metric as part of our contribution.


%
%
%
%
%

\section{Method}
\label{sec:methods}

\begin{figure*}[t]
  \centering  \includegraphics[width=\linewidth]{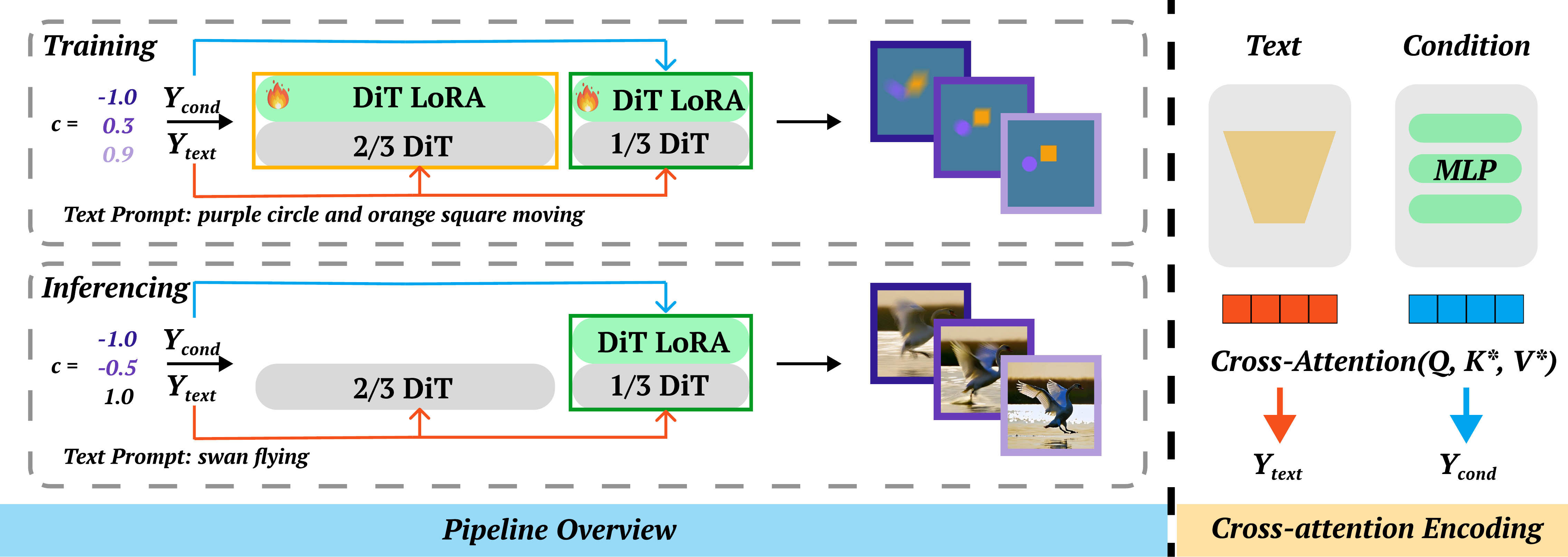}
  \caption{Overview of our controllable generation pipeline. To achieve decoupled control, we encode the scalar condition separately from the text guidance via a parallel cross-attention module. During training (top), we optimize the conditional adapter while actively updating the backbone by injecting LoRA layers into all DiT blocks. During inference (bottom), we discard the LoRA weights from the shallow two-thirds of the transformer blocks, retaining only the conditional adapter and backbone LoRA in the deepest third of the blocks. This selective retention enables high-fidelity physical control while minimizing semantic corruption of the backbone.}
  \label{fig:pipeline}
\end{figure*}

Our primary objective is to introduce direct, low-dimensional, continuous control over a specific attribute of the output produced by a T2V generative model. We achieve this by fine-tuning a pretrained T2V model on a carefully assembled dataset, which exhibits sufficient variation along the desired control axis. To enable this fine-tuning, we introduce \textit{two architectural modifications} to the base model. Our overall training architecture is illustrated in~\Cref{fig:pipeline}, and the supplementary material contains complete pseudocode for our method (Algorithm~\ref{alg:forward_pass}).


\vspace{1mm}
\noindent\textbf{Disentangled Conditioning Module.}
Following recent work in controllable generation \cite{ye2023ip}, we employ a dedicated cross-attention module to inject the conditioning signal. We normalize the condition (which in our experiments is a scalar value) to obtain $c \in [-1, 1]$. This condition is projected to a higher-dimensional embedding vector via a small multi-layer perceptron (MLP): $e_{\text{cond}} = \operatorname{MLP}_{\text{cond}}(c)$. The embedded conditional signal $e_{\text{cond}}$ is then injected into the model using a dedicated cross-attention mechanism that operates in parallel with the main text cross-attention. We insert this conditional cross-attention adapter into the deepest third of the original model's transformer blocks. For a given query $q$ from the video latent, the output of the cross-attention block is a linear combination of the text-conditioning $y_{\text{text}}$ and the attribute-conditioning $y_{\text{cond}}$.


\noindent\textbf{Backbone LoRA.}
\label{sec:method_adapter}
Rather than attempting to construct a dataset for fine-tuning that uniformly and consistently exhibits variation along the desired control axis \emph{and} matches the natural video distribution of the original T2V model, we make the deliberate choice to construct datasets that satisfy only the former requirement.
This decision frees us from carefully curating natural videos or performing difficult photorealistic rendering and instead allows us to focus on creating simple synthetic examples that illustrate the conditioning effect. However, fine-tuning on such out-of-domain datasets inevitably induces content drift. To address this, we employ a joint training strategy where a backbone LoRA is optimized alongside the conditioning module. This approach enforces a ``separation of concerns'': the backbone LoRA absorbs the synthetic domain shift, allowing the conditional adapter to focus exclusively on disentangling the physical control signal. Critically, at inference time, we revert this drift by selectively discarding the backbone LoRA weights from all blocks that are not equipped with the conditional adapter. This restores the model's original generative priors in the majority of the network while preserving the learned control mechanism.


\section{Dataset Construction}
\label{sec:dataset}



We evaluate our method on three scalar camera parameters—shutter speed, aperture, and color temperature—using synthetic datasets that isolate their respective effects: motion blur, depth of field, and color tone. We adopt a simple, non-photorealistic domain to enable precise procedural control and clean variations along each conditioning axis. To ensure diversity within this controlled setting, we apply two forms of scene randomization.

\vspace{1mm}
\noindent\textbf{Control Scalar Randomization.}
The primary driver of generalization in our method is the low-dimensional adapter (\cref{sec:method_adapter}). To complement this and ensure a continuous response, we employ a multi-level stratified sampling strategy rather than relying on fixed discrete values.
We partition the normalized range $[-1, 1]$ into $N$ equal-width bins and sample uniformly from within each bin, applying this strategy across several hierarchical layers. This ``pyramid'' approach ensures that our sparse dataset provides a rich, continuous signal by concentrating sampling density near the center of the control space, preventing overfitting to specific values (see the supplementary material, \Cref{sec:dataset_details}, for full details).

\vspace{1mm}
\noindent\textbf{Scene Content Randomization.}
We design a dataset of geometric primitives that balances semantic simplicity with functional expressiveness, avoiding the confounding complexity of photorealistic data. Each training scene is procedurally generated with a randomized composition of moving shapes.
This configuration ensures that the necessary conditions for physical control are visually observable---such as varying trajectories for motion blur or overlapping depth planes for aperture---while eliminating unnecessary semantic details that could induce backbone corruption.

\vspace{1mm}
\noindent\textbf{Photorealistic Ablation Dataset.}
To test our ``Less is More'' hypothesis, we conduct a controlled ablation comparing data content. We construct a high-fidelity ``photorealistic" dataset and a ``simple synthetic" counterpart. Each dataset consists of only one scene per physical effect, thereby isolating data complexity as the sole experimental variable.
The photorealistic scenes, characterized by high semantic complexity, are as follows:
\begin{itemize}
    \item \textbf{Shutter Speed:} We simulate motion blur by applying temporal averaging (frame accumulation) to clips from a high-speed (240-fps) video of a dynamic scene.
    \item \textbf{Aperture:} We generate depth-of-field variations by applying BokehMe~\cite{Peng2022BokehMe} to a single image featuring a complex foreground subject and detailed background.
    \item \textbf{Color Temperature:} We generate color variations by applying a physics-based Kelvin transformation to a single natural outdoor scene with a broad color palette.
\end{itemize}
This dataset is used in our experiments (\cref{sec:experiments}) to demonstrate the risks of backbone corruption.
\section{Evaluation Methodology}
\label{sec:quan_evalulation}

We propose a two-stage evaluation framework to measure the drift and corruption of the model quantitatively. The first stage, our novel Fast Evaluation Protocol (FEP), introduces lightweight metrics to quantify the magnitude and velocity of backbone changes during fine-tuning. The second stage, the Slow Validation Protocol (SVP), employs established, full-scale metrics to assess the final generative quality and temporal coherence, allowing us to interpret the nature of the changes measured by FEP.

\subsection{Stage 1: Fast Evaluation Protocol (FEP)}
\label{sec:fep}

The FEP is a computationally efficient protocol designed to quantify the magnitude of semantic and distributional shifts in a fine-tuning model relative to its original state. The protocol consists of a single-step denoising pass to generate a minimal 4-frame output from a fixed latent seed. This pass is repeated across a broad, semantically diverse prompt set, with specific details provided in our experimental setup (\cref{sec:experiments}).

\vspace{1mm}
\noindent\textbf{FEP Metrics.} We define two metrics that compare the single-step output of an adapted checkpoint against the original (pre-trained) backbone in the CLIP embedding space.

\begin{itemize}
    \item \textbf{Single-Step Fidelity (SSF) Score:} The average per-prompt cosine similarity between the original backbone's and the adapted checkpoint's embeddings. This measures adherence to the original model's semantic priors. A score near 1.0 indicates minimal semantic change, while a low score signifies a major shift.

    \item \textbf{Single-Step Fréchet Distance (SS-FD):} The Fréchet Distance between the full embedding distributions of the original backbone and the adapted checkpoint. This measures holistic distributional divergence. A high score indicates a significant shift in the model's output distribution away from the original.
\end{itemize}

\noindent To establish a baseline for inherent variance, we compute these metrics by comparing the original T2V backbone against itself using only different random seeds. This baseline captures minimal, non-deterministic drift. An adapted checkpoint is thus considered to have minimal backbone change if its SSF and SS-FD scores are comparable to this baseline value.

\vspace{1mm}
\noindent\textbf{Quantifying Dataset Complexity via Drift Rate.}
The FEP metrics enable us to quantify dataset complexity by its impact on the model. We define the Distributional Drift Rate ($\mathcal{V}_{\text{drift}}$) as the rate of change in our FEP metrics: $\mathcal{V}_{\text{drift}} = \delta(\text{SS-FD}) / \delta(\text{steps})$. A low-complexity dataset (e.g., our 2D primitives) exhibits a low $\mathcal{V}_{\text{drift}}$. Conversely, a high-complexity dataset (e.g., photorealistic scenes) induces a high $\mathcal{V}_{\text{drift}}$. This metric thus provides a quantitative measure of dataset complexity by reflecting the magnitude of change it induces in the backbone.

\subsection{Stage 2: Slow Validation Protocol (SVP)}
\label{sec:svp}

The SVP stage is designed to assess the final generative quality of a model using established, full-scale metrics. It involves a full denoising process to assess both semantic fidelity and perceptual quality.

\vspace{1mm}
\noindent\textbf{SVP Metrics.} We assess two primary categories of metrics:

\begin{itemize}
    \item \textbf{Semantic Fidelity (Backbone Health):} We compute video-text alignment using the X-CLIP Score \cite{ni2022expanding} and the VQA score \cite{lin2024evaluating}. Cross-referencing these scores with the $\mathcal{V}_{\text{drift}}$ from FEP allows us to diagnose the adaptation: a high-velocity drift is \textit{benign} if final scores are high (successful adaptation), but \textit{malignant} if they collapse (catastrophic forgetting).

    \item \textbf{Video Quality \& Temporal Coherence:} We use six metrics from VBench \cite{huang2024vbench} to evaluate the generated video quality independent of text guidance: subject/background consistency, motion smoothness, dynamic degree, aesthetic quality, and imaging quality.
\end{itemize}

Together, FEP and SVP form a complementary two-stage framework. FEP provides frequent, low-cost monitoring of backbone drift during training---its CLIP-based single-step measurements are intentionally lightweight, serving as early-warning signals for overfitting (e.g., prompt degradation or training-context copying) that correlate with downstream SVP degradation. Full temporal coherence and video fidelity, which single-step CLIP proxies are not designed to capture, are assessed by SVP through complete multi-step denoising. This division of responsibility enables efficient training-time diagnosis without the computational overhead of full video generation at every checkpoint.


\begin{table*}[t]
  \centering
  \caption{
    \textbf{Quantitative SVP Results.}
    Comparison of \textbf{Full-LoRA} (Full) and \textbf{Decoupled} (Dec.) inference against the original backbone baseline across three controls. Within each control pair, \colorbox{hlcell}{highlighted} values are closer to the baseline.
  }
  \label{tab:svp_results}
  \newcommand{\hl}[1]{\colorbox{hlcell}{#1}}
  \begin{tabular*}{\textwidth}{@{\extracolsep{\fill}}l c cc cc cc@{}}
    \toprule
    & & \multicolumn{2}{c}{\textbf{Shutter (ST)}} & \multicolumn{2}{c}{\textbf{Aperture (AP)}} & \multicolumn{2}{c}{\textbf{Temperature (TP)}} \\
    \cmidrule(lr){3-4} \cmidrule(lr){5-6} \cmidrule(lr){7-8}
    \textbf{Metric} & \textbf{Base.} & Full & Dec. & Full & Dec. & Full & Dec. \\
    \midrule
    \multicolumn{8}{l}{\textit{Semantic Fidelity}} \\
    \quad X-CLIP Score & 25.390 & \hl{25.295} & 25.587 & 25.181 & \hl{25.595} & \hl{25.487} & 25.595 \\
    \quad VQA Score & 0.522 & 0.453 & \hl{0.521} & 0.427 & \hl{0.513} & 0.550 & \hl{0.532} \\
    \midrule
    \multicolumn{8}{l}{\textit{Video Quality}} \\
    \quad Subject Consistency & 0.951 & 0.939 & \hl{0.946} & 0.968 & \hl{0.951} & 0.960 & \hl{0.950} \\
    \quad Background Consistency & 0.973 & 0.963 & \hl{0.969} & 0.979 & \hl{0.973} & 0.975 & \hl{0.972} \\
    \quad Motion Smoothness & 0.988 & 0.983 & \hl{0.987} & 0.994 & \hl{0.987} & 0.990 & \hl{0.989} \\
    \quad Dynamic Degree & 0.427 & 0.688 & \hl{0.469} & 0.219 & \hl{0.438} & 0.406 & \hl{0.417} \\
    \quad Aesthetic Quality & 0.606 & 0.576 & \hl{0.604} & 0.629 & \hl{0.610} & 0.620 & \hl{0.605} \\
    \quad Imaging Quality & 0.618 & 0.531 & \hl{0.596} & 0.596 & \hl{0.633} & 0.664 & \hl{0.623} \\
    \bottomrule
  \end{tabular*}
\end{table*}

\begin{figure}[t]
  \centering
  \includegraphics[width=\linewidth]{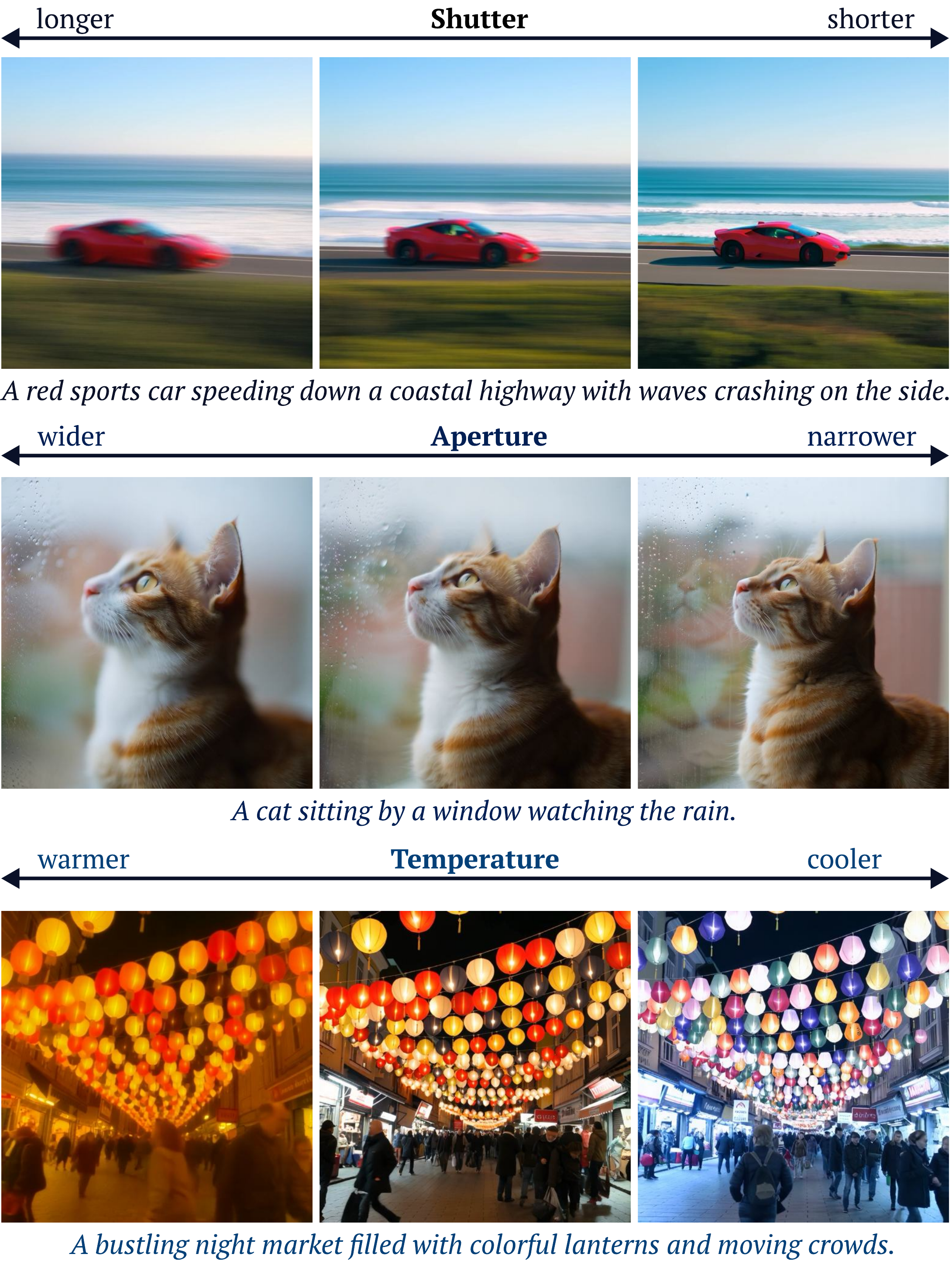}
  \caption{Qualitative results for Full-LoRA Inference. Retaining all backbone LoRA weights yields high-quality generation, supporting the robustness of our data-efficient training.}
  \label{fig:backbone_infer}
\end{figure}

\section{Experiments}
\label{sec:experiments}

To validate our ``Less is More'' hypothesis and analyze how different data strategies impact backbone fidelity, we conduct two primary groups of experiments. We use WAN 2.1 \cite{wan2025wan} as our T2V backbone for all fine-tuning and evaluation.

\subsection{Experimental Design}

\begin{itemize}
    \item \textbf{Group 1: Data Complexity (Syn vs. Real).} 
   To isolate the effect of data content, we conduct a controlled one-shot comparison. For each camera parameter, we train two models on minimal data: (1) a model trained on a single synthetic scene and (2) a model trained on a single photorealistic scene (per \cref{sec:dataset}). From each scene, we sample 7 scalar conditions to create the training data. This ablation is evaluated using our FEP on the 800-prompt VBench benchmark \cite{huang2024vbench}.

    \item \textbf{Group 2: Inference Strategy (Decoupled vs. Full-LoRA).}
    To evaluate our proposed inference strategy, we use our full model trained on the complete pyramid synthetic dataset (\cref{sec:dataset}) for each camera parameter and compare two inference modes: \textbf{Decoupled Inference} (pruning shallow DiT LoRA blocks to restore the pre-trained backbone's priors) and \textbf{Full-LoRA Inference} (retaining all backbone LoRA weights at test time). This comparison is evaluated using our SVP on a curated 96-prompt high-motion suite, which involves a full denoising process of 50 steps to generate 49 frames.
\end{itemize}

\subsection{Results and Analysis}

Our quantitative results are presented in \cref{fig:quan-compare} and \cref{tab:svp_results}.
\subsubsection{Group 1: Synthetic Data Prevents Forgetting}

\noindent\textbf{Drift Monitoring (FEP):} The results (\cref{fig:quan-compare}, Top Row Left) strongly support our ``Less is More'' hypothesis. The FEP monitoring shows that the model trained on photorealistic ``Real'' data (dashed line) exhibits a much faster rate of change in its FEP scores per training step compared to the model trained on ``Syn'' data (solid line). This observation validates that complex, photorealistic data induces a significantly faster and more severe contextual drift in the T2V backbone than our simple, synthetic data.

\noindent\textbf{Corruption Analysis (SVP):} The SVP validation (\cref{fig:quan-compare}, Top Row Right) confirms the nature of this drift. The ``Real'' model's semantic scores (X-CLIP, VQA) collapse, indicating malignant, catastrophic forgetting. Conversely, our ``Syn'' model's scores remain on par with the baseline, proving simple synthetic data is superior for avoiding backbone corruption.

\subsubsection{Group 2: Decoupled Inference Retains Fidelity}

\begin{figure*}[htbp]
  \centering
  \begin{subfigure}{0.24\linewidth}
    \includegraphics[width=1.0 \linewidth]{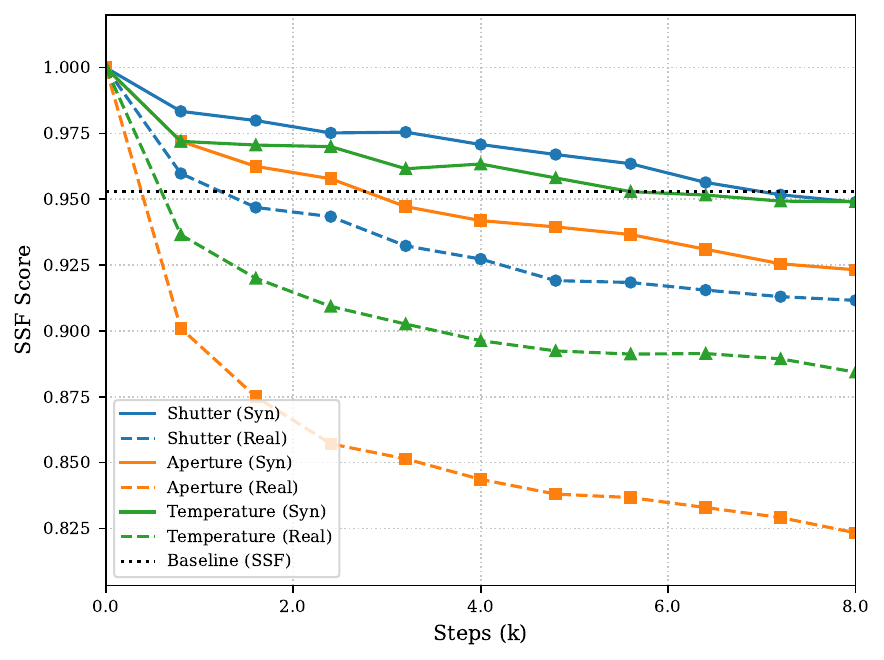}
    \label{fig:line_ssf_sr}
  \end{subfigure}
  \hfill
    \begin{subfigure}{0.24\linewidth}
    \includegraphics[width=1.0 \linewidth]{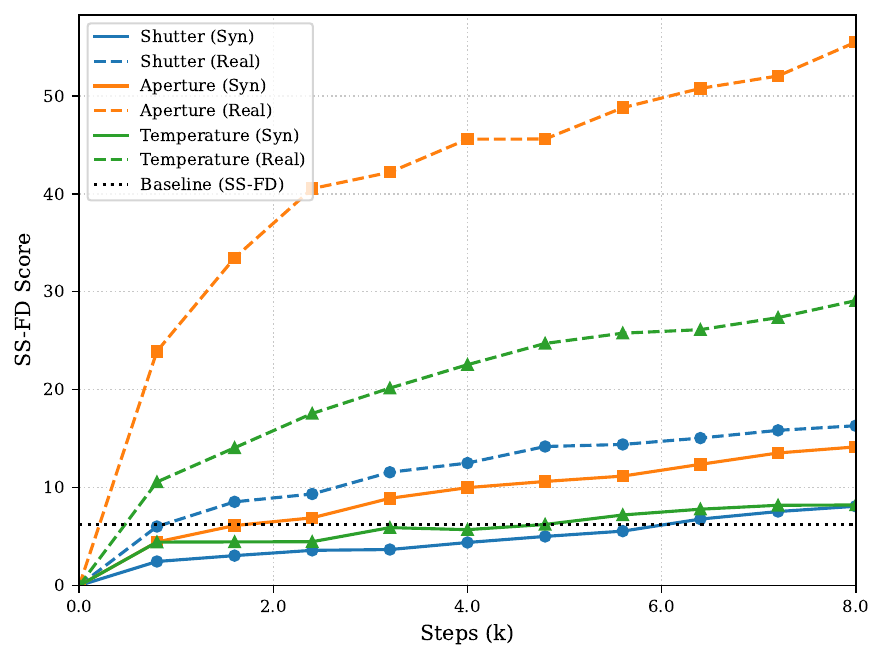}
    \label{fig:line_ssfd_sr}
  \end{subfigure}
    \begin{subfigure}{0.24\linewidth}
    \includegraphics[width=1.0 \linewidth]{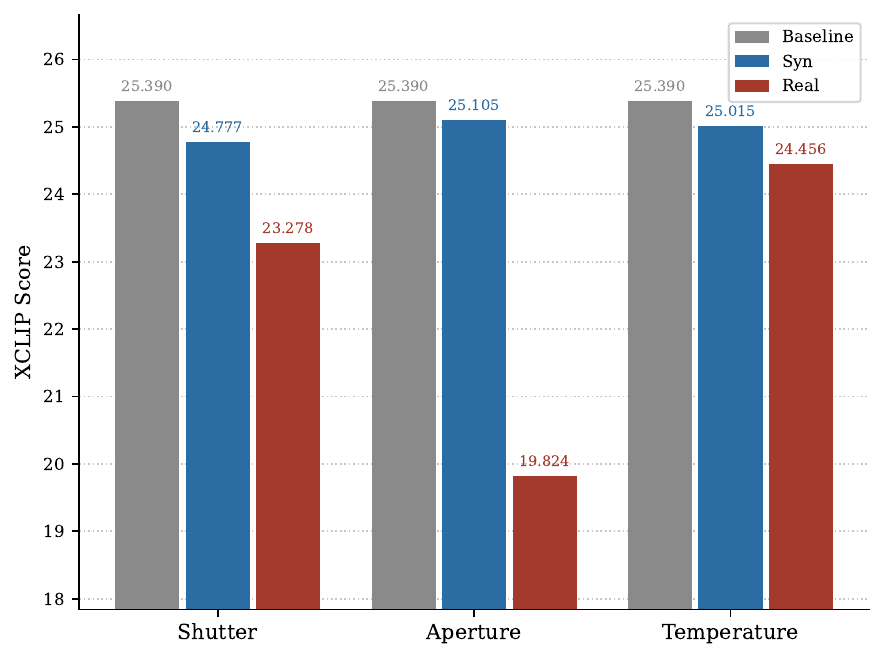}
    \label{fig:xclip_bar_sr}
  \end{subfigure}
  \hfill
    \begin{subfigure}{0.24\linewidth}
    \includegraphics[width=1.0 \linewidth]{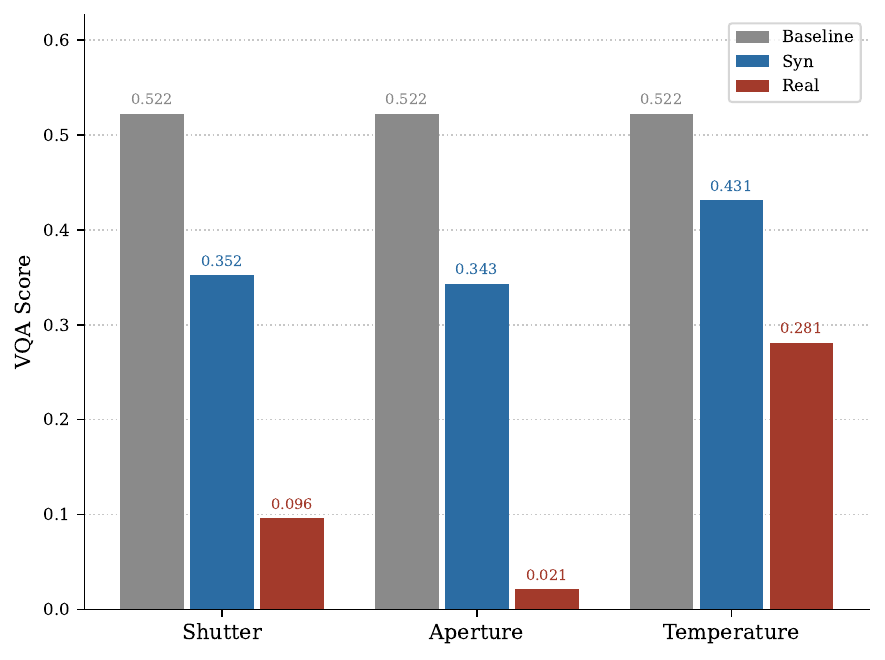}
    \label{fig:vqa_bar_sr}
  \end{subfigure}

\begin{subfigure}{0.24\linewidth}
    \includegraphics[width=1.0 \linewidth]{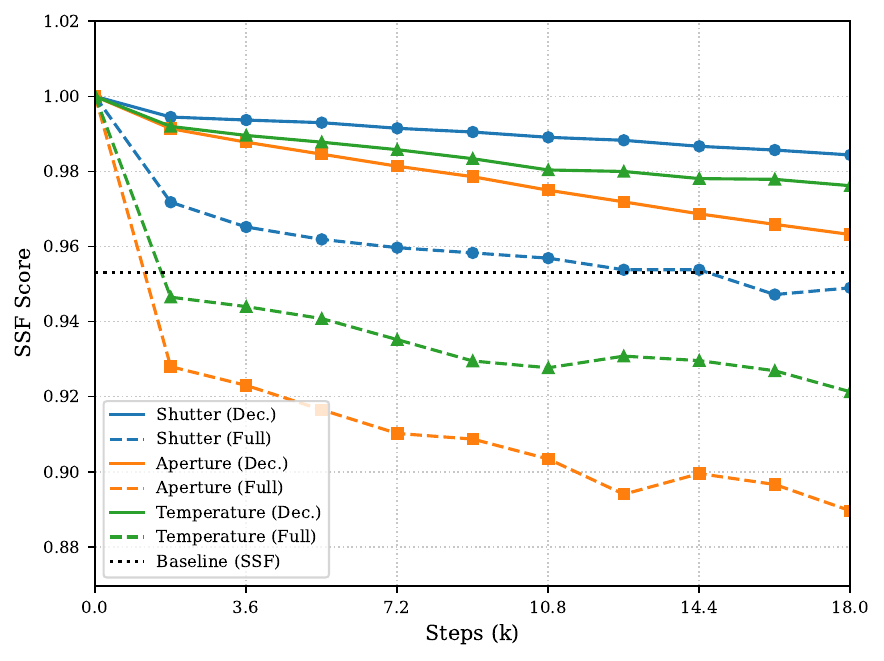}
    \label{fig:line_ssf_cd}
  \end{subfigure}
  \hfill
    \begin{subfigure}{0.24\linewidth}
    \includegraphics[width=1.0 \linewidth]{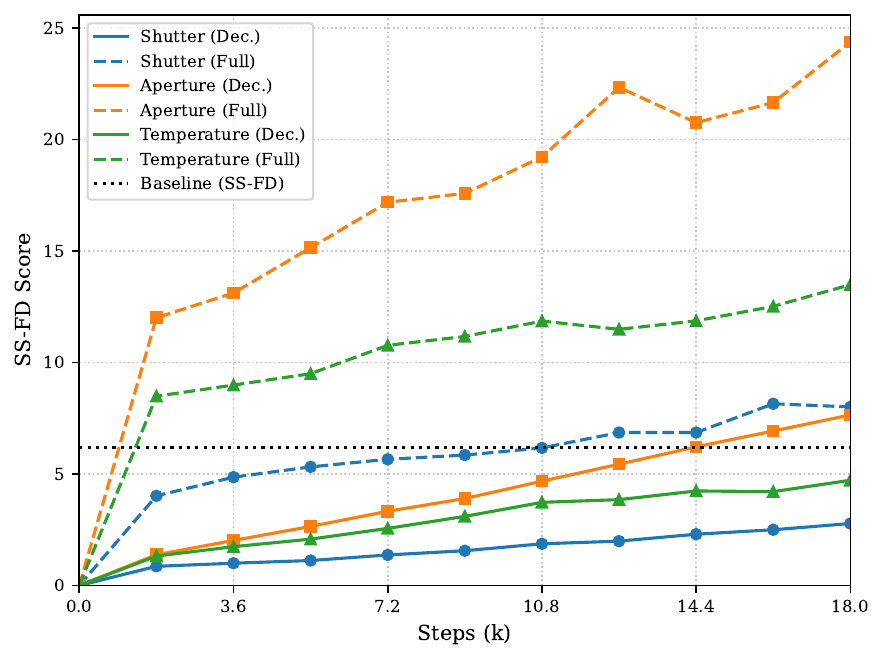}
    \label{fig:line_ssfd_cd}
  \end{subfigure}
    \begin{subfigure}{0.24\linewidth}
    \includegraphics[width=1.0 \linewidth]{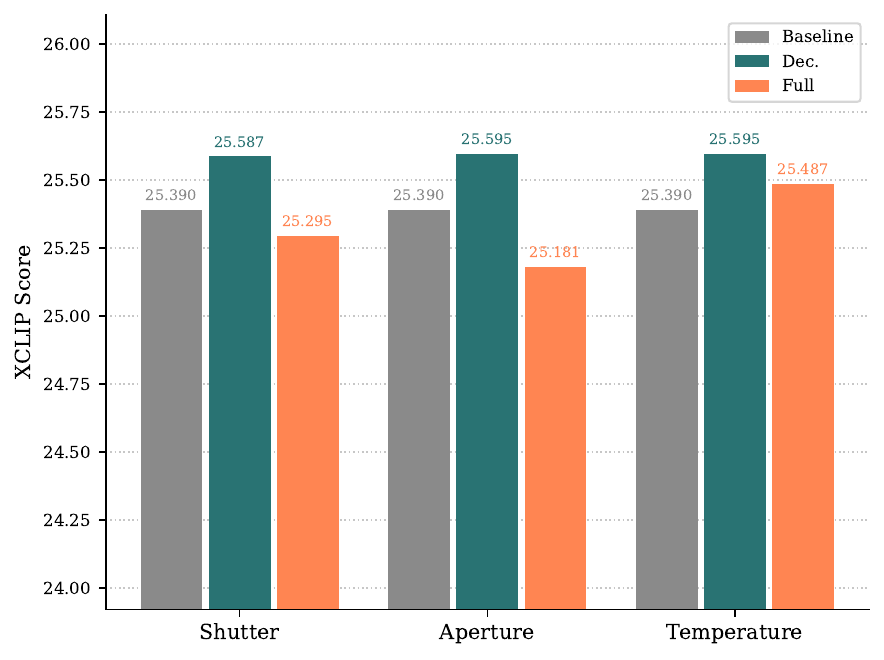}
    \label{fig:xclip_bar_cd}
  \end{subfigure}
  \hfill
    \begin{subfigure}{0.24\linewidth}
    \includegraphics[width=1.0 \linewidth]{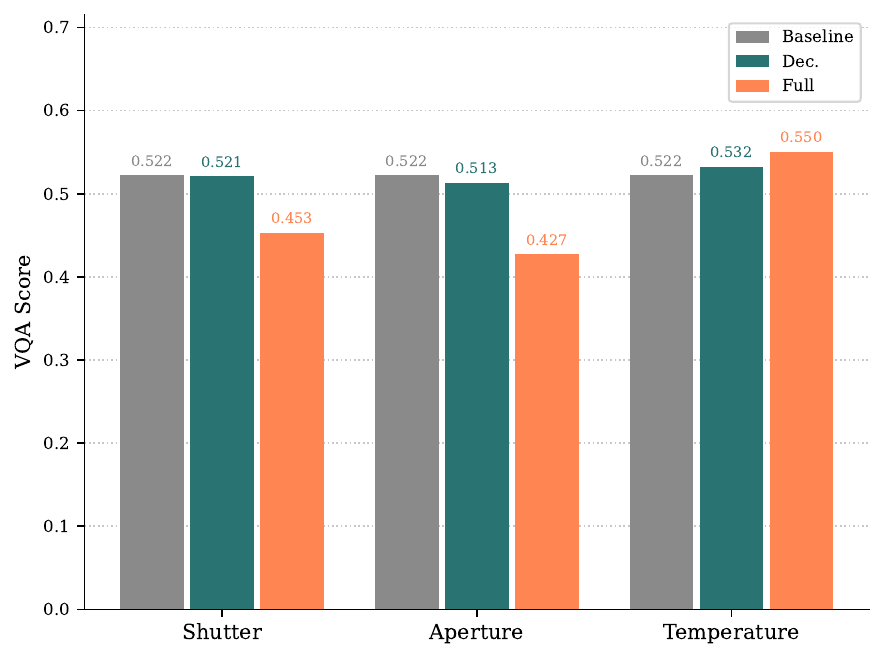}
    \label{fig:vqa_bar_cd}
  \end{subfigure}
  
\caption{\textbf{Ablation studies on data complexity and inference strategy.}
\textbf{Top Row}: Synthetic vs. Real Data. A one-shot comparison of fine-tuning on our low-fidelity synthetic (``Syn'') versus complex photorealistic (``Real'') data. (Left) FEP monitoring tracks SSF and SS-FD over training steps. (Right) SVP validation bar charts show final X-CLIP and VQA scores.
\textbf{Bottom Row}: Decoupled vs. Full-LoRA Inference. An ablation on our full pyramid-trained model. (Left) FEP monitoring tracks SSF and SS-FD against training steps. (Right) SVP validation bar charts compare final X-CLIP and VQA scores for Decoupled Inference and Full-LoRA Inference.}
\label{fig:quan-compare}
\end{figure*}

\noindent\textbf{FEP \& SVP Analysis:} The results (\cref{fig:quan-compare}, Bottom Row Right) validate the superiority of our proposed \textbf{Decoupled Inference}. The FEP analysis shows it barely impacts the backbone, with drift remaining around the baseline. The SVP analysis confirms this, showing minimal semantic corruption (SVP scores change $<$2\%). Meanwhile, \textbf{Full-LoRA Inference}, while inferior, shows only a slight, acceptable score reduction. This confirms our training is robust and does not cause catastrophic degradation, remaining capable of high-fidelity controllable generation (\cref{fig:backbone_infer}).

\noindent\textbf{Video Quality (VBench):} As shown in \cref{tab:svp_results}, a key finding is the stability of the VBench video quality metrics. Both Decoupled and Full-LoRA Inference on models trained on the synthetic dataset maintain scores for Subject Consistency, Background Consistency, Motion Smoothness, and Aesthetic Quality that are nearly identical to the original baseline. This is a crucial result: it demonstrates that our dataset selection and inference design successfully disentangle the new physical control without introducing unwanted artifacts or degrading the backbone's core generative quality, which is a common failure case.

\subsection{Visual Condition Comparison}
\begin{figure*}[ht]
  \centering
  \includegraphics[width=\linewidth]{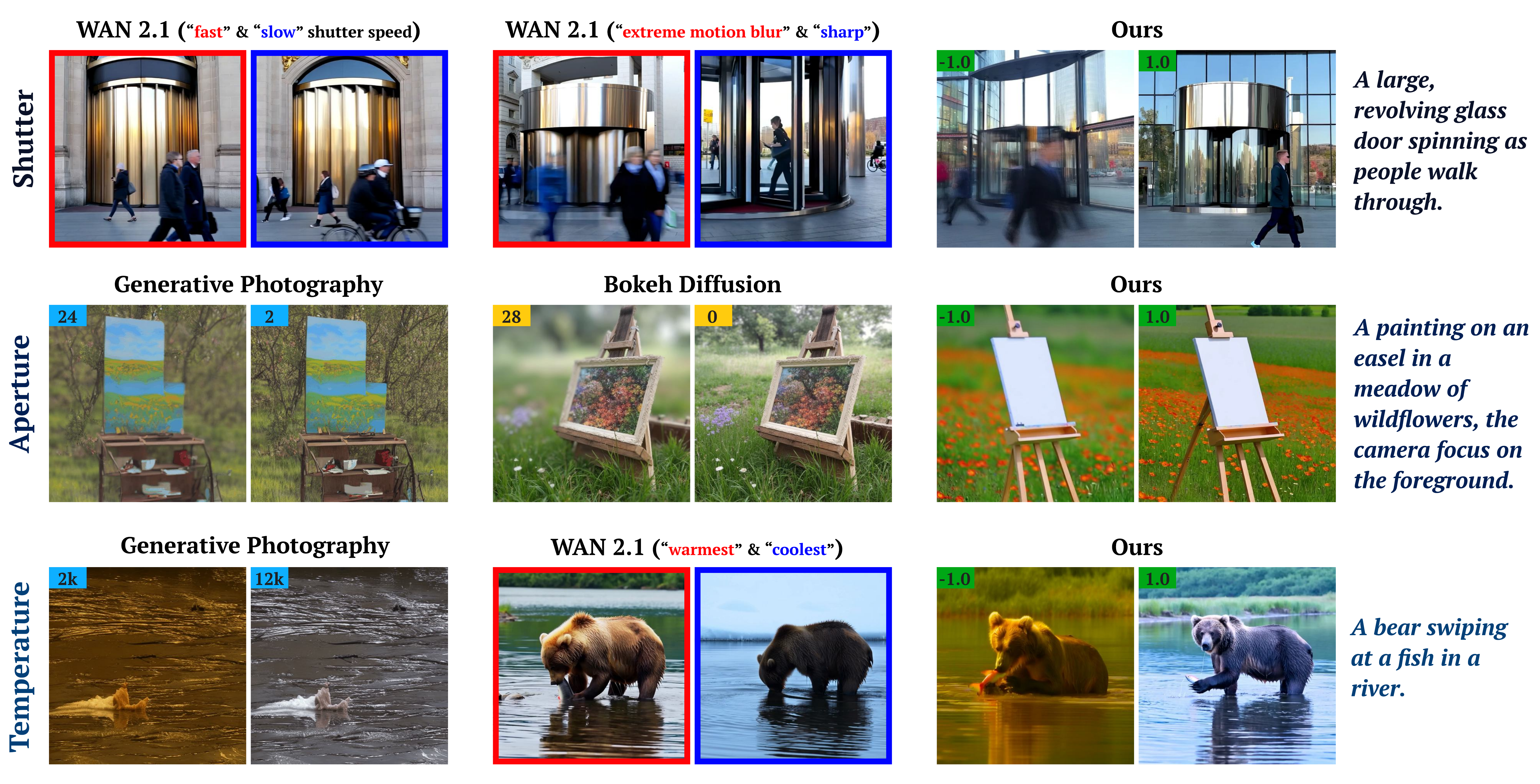}
  \caption{Qualitative comparison of physical controls. We compare our method against text-based prompting in a T2V backbone (WAN 2.1) and specialized, data-intensive image baselines Generative Photography~\cite{Yuan_2024_GenPhoto} and Bokeh Diffusion~\cite{fortes2025bokeh}.}
  \label{fig:compare}
\end{figure*}

Apart from the comparisons via training data and inference methods, we also conduct qualitative comparisons against WAN 2.1~\cite{wan2025wan}, Bokeh Diffusion~\cite{fortes2025bokeh}, and Generative Photography~\cite{Yuan_2024_GenPhoto}.
\cref{fig:compare} compares the conditioning ability of different workflows. Text-based prompting in the T2V backbone (WAN 2.1) exhibits significant semantic entanglement.
This is particularly evident in the Shutter Speed test (Row 1), where the baseline either struggles to associate shutter speed changes with motion blur, or generates a lower-fidelity scene by creating a ``metallic'' door instead of the requested glass one.
A similar entanglement occurs in the Temperature test (Row 3), where the baseline misinterprets color temperature as weather, incorrectly altering the scene's environment (e.g., adding snow) rather than its color tone.
Our method overcomes these shortcomings by correctly applying the physical effect of shutter speed (motion blur) without this semantic or fidelity confusion.
Furthermore, for Aperture (Row 2), our model generates high-fidelity bokeh effects comparable to specialized, data-intensive baselines such as Bokeh Diffusion \cite{fortes2025bokeh} and Generative Photography \cite{Yuan_2024_GenPhoto}.
These results validate our central hypothesis: our data-efficient synthetic training achieves robust physical disentanglement via a low-rank adapter, eluding the limitations of entangled text prompts and rivaling the quality of data-heavy methods.
The supplementary material (see \Cref{sec:gallery}) provides a number of additional examples (as videos) of our method that demonstrate the consistency on a range of control values.

\section{Analysis}
\label{sec:qual_eva}
Beyond our quantitative metrics (\cref{sec:quan_evalulation}), we develop a suite of data-free analysis techniques. These methods directly analyze the geometric properties of the model's weights and their functional outputs, removing any dependency on specific data samples at test time. To validate our hypotheses on adaptation and disentanglement, we briefly introduce these techniques here and defer full implementation details to the supplementary material.


\subsection{Hypothesis 1: Low-Fidelity Data Prevents Catastrophic Forgetting}

Our first hypothesis is that fine-tuning on low-fidelity synthetic data results in a cleaner backbone LoRA update, preserving the pre-trained model's general knowledge. To validate this, we apply the ``intruder dimension'' analysis from~\citet{shuttleworth2025loravsfinetuningillusion}: we perform SVD on the original backbone weight $W_{\text{pre}}$ and its adapted version $W_{\text{lora}} = W_{\text{pre}} + \Delta W_{\text{lora}}$, defining an intruder as a new, high-ranking singular vector in $W_{\text{lora}}$ with cosine similarity $< \epsilon$ to all vectors in $W_{\text{pre}}$. A high intruder count is a mathematical signature of catastrophic forgetting; a low count confirms a non-destructive adaptation.

The full results are provided in the supplementary material (Algorithm~\ref{alg:backbone_drift}, \Cref{sec:heatmap}): synthetic training produces minimal backbone drift while real-data training induces substantial high-ranking intruder dimensions---independent spectral evidence for our ``Less is More'' hypothesis.

\subsection{Hypothesis 2: Low-Rank Conditional Representation}
Our second hypothesis is that the adapter has learned an efficient, compact representation for the physical effect---a sign of good generalization. We test this via an effective rank analysis: we generate $V_{\text{cond}}(c)$ for a strong condition ($c=1$), reshape it into a 2D matrix, and compute the SVD of the resulting conditional output $y_{\text{cond}}$ (see Algorithm~\ref{alg:evaluation} for full details). A physical effect such as motion blur should not require a high-dimensional feature space; a low-rank singular value spectrum therefore indicates that the adapter has learned the essence of the effect rather than overfitting to training data artifacts. Extended visual comparisons across four training configurations are provided in the supplementary material (\Cref{sec:low_rank_vis}).

\begin{figure}[t!]
  \centering
  \begin{subfigure}{0.48\linewidth}
    \includegraphics[width=1.0 \linewidth]    {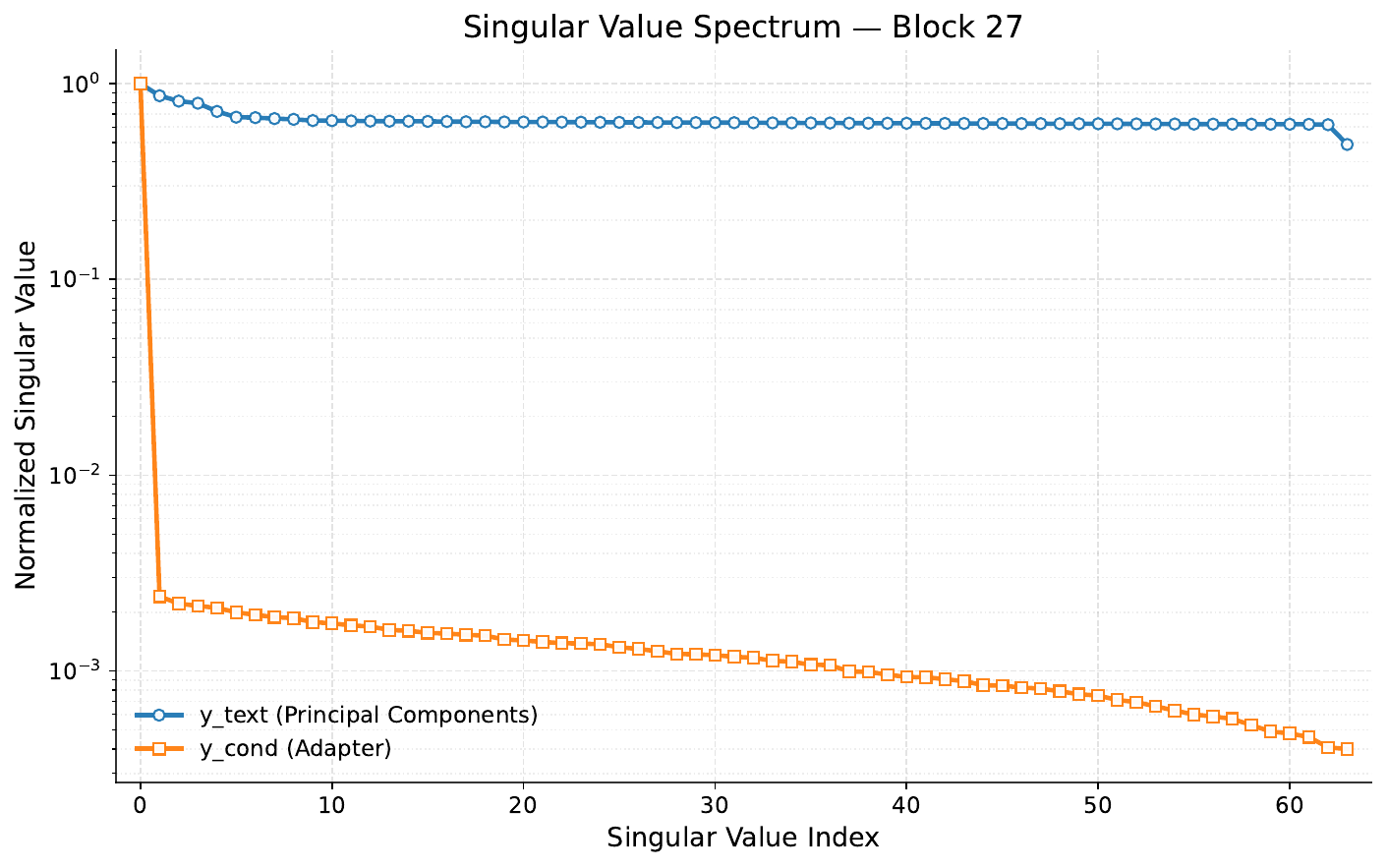}
    \caption{Joint Training}
    \label{fig:orthogonal_good_block27}
  \end{subfigure}
  \hfill
  \begin{subfigure}{0.48\linewidth}
    \includegraphics[width=1.0 \linewidth]
    {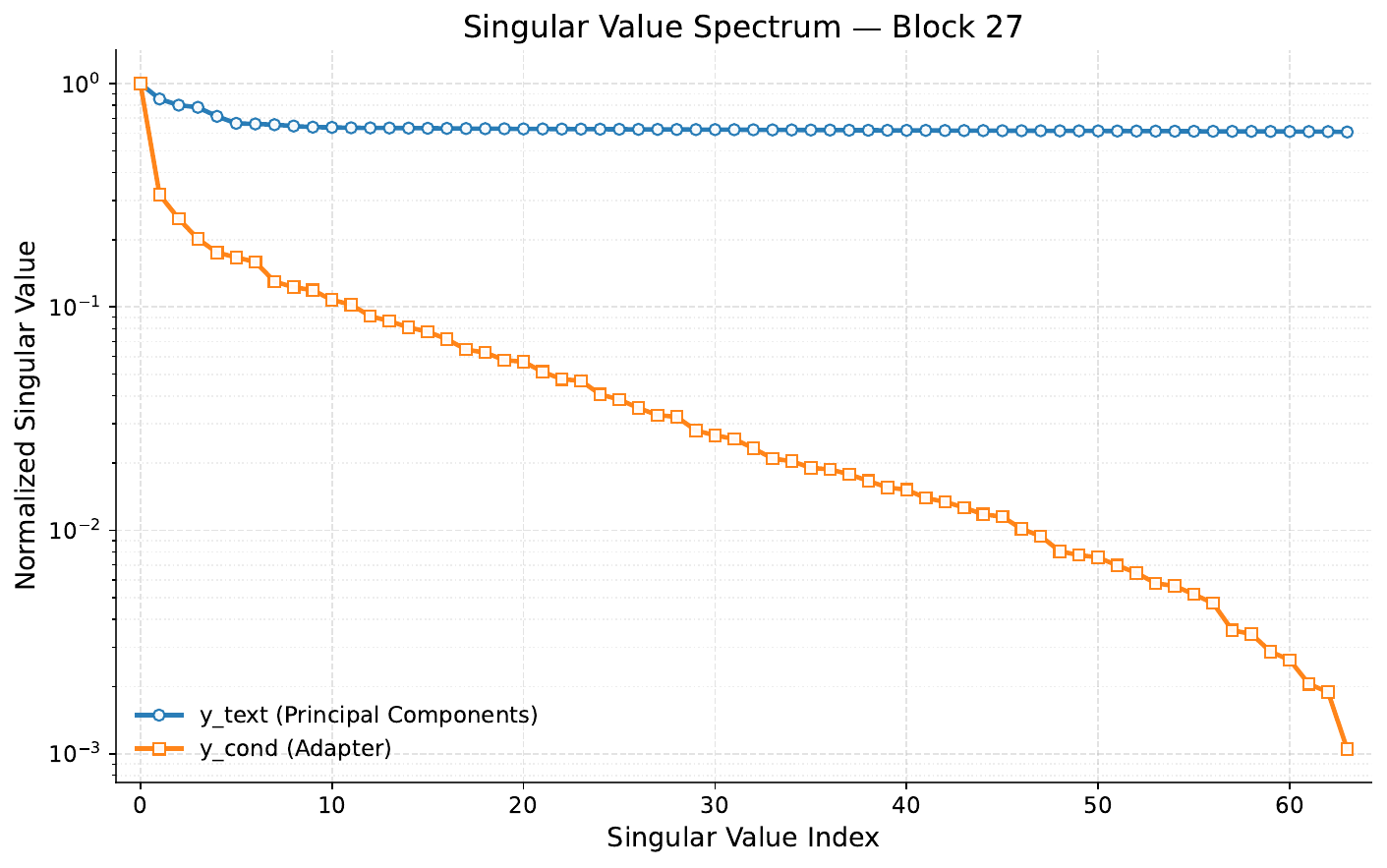}
    \caption{Adapter-Only Training}
    \label{fig:orthogonal_bad_block27}
  \end{subfigure}
  \caption{\textbf{Singular value spectrum of the conditional signal $y_{\text{cond}}$ in Block~27.} (a)~In our jointly trained model, the conditional signal exhibits a sharp spectral decay with an effective rank of~1, confirming that the adapter learned an efficient, low-dimensional representation of the physical effect. (b)~In the adapter-only model, the signal is high-rank, with a slow spectral decay that mirrors the content signal $y_{\text{text}}$, indicating that the adapter memorized the training data content rather than isolating the conditioning effect.}
  \label{fig:rank_comparison}
\end{figure}

\noindent\textbf{Jointly Trained Checkpoint.} The jointly trained model demonstrates successful factorization. We analyze the singular value spectrum of its conditional signal $y_{\text{cond}}$ (\cref{fig:orthogonal_good_block27}), which reveals a classic ``elbow'' curve with an effective rank of~1. This result confirms the adapter learned an efficient, low-dimensional representation of the \emph{effect}, as the base LoRA handled the domain's content. This low-rank structure is the key to generalization.

\noindent\textbf{Adapter-Only Checkpoint.} In stark contrast, the adapter-only model fails. Its conditional signal $y_{\text{cond}}$ has a high-rank spectrum that decays slowly, mirroring the content signal $y_{\text{text}}$ (\cref{fig:orthogonal_bad_block27}). This is the root cause of its failure: trained in isolation, the adapter did not learn the simple effect but instead memorized the complex, high-dimensional content of the synthetic training data.

\noindent\textbf{The ``Bulldozer Effect''.} We term this failure mode the ``Bulldozer Effect''. The adapter-only signal is not only \textbf{high-rank} (content memorization) but also has a substantially larger magnitude and is orthogonal to the backbone's content space. Instead of collaborating, the adapter represents its memorized content in a vacant subspace. During inference, this powerful signal ``bulldozes'' and replaces the text signal $y_{\text{text}}$, resulting in output dominated by the training scene features.

\section{Conclusion and Future Work}
As increasingly powerful pre-trained generative video models emerge, post-training strategies will become all the more crucial to achieve reliable, intuitive, and controllable models for specific use cases.
We offer a new paradigm for constructing fine-tuning datasets for video models.
We show that rather than attempting to gather a dataset that is as \emph{realistic} as possible, a better strategy may be to construct one that is as \emph{disentangled} as possible, foregoing realism entirely.
We posit that the latent representations learned by foundation models are so expressive and the priors for realism so strong that the role of fine-tuning is coaxing out existing learned properties within the latent space, rather than extrapolating and expanding the domain.

While we demonstrate the efficacy of our method on three specific control types, the general framework inspires many directions for future work, such as a unified model that disentangles multiple parameters (e.g., aperture \textit{and} temperature) from a single joint control vector.
Furthermore, this data-efficient paradigm could be extended to other spatial controls, such as learning depth or pose from simple geometric data, potentially reducing the need for complex, photorealistic datasets in those domains as well.

\clearpage
\setcounter{page}{1}
\appendix
\maketitlesupplementary

\section{Dataset Details}
\label{sec:dataset_details}
\textbf{Pyramid Sampling Strategy.}
We first describe our efficient sampling strategy, which balances coverage of the control space with the cost of generating synthetic scenes (see also Algorithm~\ref{alg:pyramid} in \Cref{sec:algo_details} for the corresponding pseudocode). Enumerating all combinations of control scalars and randomized scene settings would lead to a combinatorial explosion, so we adopt a \emph{pyramid generation strategy} that varies the number of sampled scalars per layer while keeping the sampling distribution uniform.

Specifically, we divide the control range $[-1,1]$ into five layers and sample $9 + 7 + 5 + 3 + 1 = 25$ scalar values using jittered sampling, ensuring that \emph{every point in the interval has equal probability of being selected}. For each layer, we generate six scenes using our randomized scene-generation procedure (described later in this section), and each scene is rendered across all scalar values assigned to its layer.

This results in $6 \times 25 = 150$ total training samples. Compared to
exhaustive sampling ($30\text{ scenes} \times 150\text{ conditions} = 4{,}500$ combinations), our pyramid strategy achieves a \textbf{30$\times$ reduction} in data size while preserving uniform coverage of the control range and diversity across the randomized context space.

\vspace{1mm}
\noindent\textbf{Shutter Speed.} 
Given a sampled scalar $c \in [-1,1]$ and a randomized scene, we synthesize videos using a simple 2D physics-based renderer. Each scene contains 1--3 colored geometric primitives (circle, square, triangle, star) moving with constant velocities on a uniform background inside a $512 \times 512$ canvas, with elastic boundary reflections.
We map $c$ to an effective frame rate $\mathrm{FPS}(c)$ via a centered log mapping, and set the exposure duration to $1/\mathrm{FPS}(c)$. For each output frame, we then integrate over the exposure window by temporally averaging 32 analytically predicted sub-frames, advancing the shapes according to their motion model at each sub-step. Crucially, the underlying shape trajectories and backgrounds are shared across all shutter values for a given scene, so the only changing factor along the control axis is the strength of the motion blur.

\vspace{1mm}
\noindent\textbf{Aperture.} 
For aperture control, we build a 3D depth-of-field image dataset in Blender. Each scene contains 2–4 rigid objects (cubes, spheres, cylinders, cones, pyramids) placed on a ground plane with a vertical back wall, both sharing a neutral or softly colored material. Objects are assigned distinct colors and arranged at different depths between $d_{\min}$ and $d_{\max}$, forming a clear foreground–midground–background ordering. We always select one foreground object as the focus target and attach the camera’s DoF focus to this object.

Given a sampled scalar $c \in [-1,1]$, we map it to a physical $f$-number using the same centered log mapping used for shutter speed, i.e., $c \mapsto \text{fstop}(c) \in [\text{fstop}_{\text{lo}}, \text{fstop}_{\text{hi}}]$. For each scene, we then render multiple conditions by varying only the aperture $\,\text{fstop}(c)\,$ while keeping camera pose, object geometry, materials, and lighting fixed. Lighting consists of a gray world background plus an optional point light placed on a circle around the scene (either per-scene or per-render), which introduces mild but controlled highlights without changing depth layering. 
This produces aligned image sets in which the foreground remains in focus while mid- and background objects exhibit depth-dependent defocus changes in accordance with the aperture control scalar.

\vspace{1mm}
\noindent\textbf{Temperature.}
We start from sharp 2D scenes and apply a 
global white-balance transformation. 
Each base scene consists of 2–4 colored 
shapes (circles, squares, triangles, stars) randomly placed on a $512\times512$ canvas with a uniform background color. We first render a single reference image per scene without blur or noise. Then, for each sampled scalar $c \in [-1,1]$, we map $c$ to a target color temperature $K(c) \in [K_{\mathrm{lo}}, K_{\mathrm{hi}}]$ using a perceptually uniform Mired-based mapping, and apply a Kelvin-to-RGB white-balance adjustment to the entire image. The transform is defined relative to a reference neutral temperature (e.g., $K_{\mathrm{ref}}=6500\,\mathrm{K}$) and optionally re-normalized to preserve average luminance. This yields condition-aligned image sets in which geometry, layout, and background remain fixed, while only the global color cast smoothly shifts from warm to cool as a function of the temperature control scalar.

\section{Algorithmic Details}
\label{sec:algo_details}

We provide high-level pseudocode for the pyramid sampling strategy used in dataset construction, our model's core attention block, and our data-free evaluation methodologies.

\subsection{Jointly Trained Attention Block}

Algorithm \ref{alg:forward_pass} outlines the forward pass of a \texttt{WanAttentionBlock} that has been adapted with our joint training strategy. It details how the base LoRA update is integrated into the backbone's weights and how the conditional signal from the \texttt{FPSCrossAttentionAdapter} is computed and combined with the text-based content signal.

\begin{algorithm}[h]
\caption{Forward Pass of a Jointly Trained Attention Block}
\label{alg:forward_pass}
\DontPrintSemicolon
\SetKwInOut{Input}{Input}
\SetKwInOut{Output}{Output}
\SetKwFunction{Attention}{Attention}
\SetKwFunction{MLP}{MLP}

\Input{
    $x$: Input video latent features\;
    $c_\text{text}$: Text context embedding\;
    $c_\text{cond}$: Scalar physical condition (e.g., in [-1, 1])\;
    $W_q, W_k, W_v, W_o$: Pre-trained cross-attention weights\;
    $\Delta W_\text{lora}$: Learned base LoRA update\;
    $\text{Adapter}_\text{cond}$: Trained conditional adapter module\;
}
\Output{
    $x_\text{out}$: Updated video latent features\;
}
\BlankLine
\tcc{1. Form the final adapted backbone weights}
$W_{q}' \leftarrow W_q + \Delta W_\text{lora\_q}$\;
$W_{k}' \leftarrow W_k + \Delta W_\text{lora\_k}$\;
$W_{v}' \leftarrow W_v + \Delta W_\text{lora\_v}$\;
$W_{o}' \leftarrow W_o + \Delta W_\text{lora\_o}$\;
\BlankLine
\tcc{2. Compute Query, Text Keys, and Text Values}
$q \leftarrow \text{norm}(x \cdot (W_{q}')^T)$\;
$K_\text{text} \leftarrow \text{norm}(c_\text{text} \cdot (W_{k}')^T)$\;
$V_\text{text} \leftarrow c_\text{text} \cdot (W_{v}')^T$\;
\BlankLine
\tcc{3. Compute the text-based content signal}
$y_\text{text} \leftarrow \Attention(q, K_\text{text}, V_\text{text})$\;
\BlankLine
\tcc{4. Compute Conditional Keys and Values via Adapter}
$e_\text{cond} \leftarrow \MLP_\text{cond}(c_\text{cond})$\;
$K_\text{cond}, V_\text{cond} \leftarrow \text{Adapter}_\text{cond}(e_\text{cond})$\;
\BlankLine
\tcc{5. Compute the conditional signal}
$y_\text{cond} \leftarrow \Attention(q, K_\text{cond}, V_\text{cond})$\;
\BlankLine
\tcc{6. Combine signals with a learned gate $g$}
$y_\text{combined} \leftarrow y_\text{text} + g \cdot y_\text{cond}$\;
\BlankLine
\tcc{7. Apply final output projection}
$x_\text{out} \leftarrow x + y_\text{combined} \cdot (W_{o}')^T$\;
\end{algorithm}

\subsection{Backbone Spectral Drift Analysis}

To support the examination of Hypothesis~1 in
\Cref{sec:qual_eva}, we evaluate how fine-tuning modifies the
backbone’s attention projections.  
Algorithm \ref{alg:backbone_drift} describes the ``Backbone Intruder Dimension Check.'' This data-free method quantifies catastrophic forgetting by measuring the spectral similarity between the original pre-trained backbone weights and the final adapted backbone weights after joint training.

\begin{algorithm}[h]
\caption{Backbone Intruder Dimension Check}
\label{alg:backbone_drift}
\DontPrintSemicolon
\SetKwInOut{Input}{Input}
\SetKwInOut{Output}{Output}
\SetKwFunction{SVD}{SVD}
\SetKwFunction{Abs}{abs}
\SetKwFunction{Max}{max}

\Input{
    $W_\text{pre}$: A pre-trained backbone weight matrix\;
    $W_\text{lora}$: The final adapted backbone weight matrix ($W_\text{pre} + \Delta W_\text{lora}$)\;
    $k$: Number of top singular vectors to check (e.g., 64)\;
    $\epsilon$: Intruder similarity threshold (e.g., 0.5)\;
}
\Output{
    $N_\text{intruders}$: Count of detected intruder dimensions\;
}
\BlankLine
\tcc{1. Compute SVD for both weight matrices}
$U_\text{pre}, \_, \_ \leftarrow \SVD(W_\text{pre})$\;
$U_\text{lora}, \_, \_ \leftarrow \SVD(W_\text{lora})$\;
\BlankLine
$N_\text{intruders} \leftarrow 0$\;
\tcc{2. Check top-k vectors of the adapted backbone}
\For{$j \leftarrow 1$ \KwTo $k$}{
    $u_\text{lora\_j} \leftarrow j$-th column of $U_\text{lora}$\;
    \BlankLine
    \tcc{3. Find max similarity to any pre-trained vector}
    $S_\text{max} \leftarrow 0$\;
    \For{$i \leftarrow 1$ \KwTo number of columns in $U_\text{pre}$}{
        $u_\text{pre\_i} \leftarrow i$-th column of $U_\text{pre}$\;
        $s \leftarrow \Abs(\text{cosine\_similarity}(u_\text{lora\_j}, u_\text{pre\_i}))$\;
        $S_\text{max} \leftarrow \Max(S_\text{max}, s)$\;
    }
    \BlankLine
    \tcc{4. Count as intruder if similarity is below threshold}
    \If{$S_\text{max} < \epsilon$}{
        $N_\text{intruders} \leftarrow N_\text{intruders} + 1$\;
    }
}
\BlankLine
\Return{$N_{\text{intruders}}$}\;
\end{algorithm}

\subsection{Data-Free Spectral Evaluation of the Conditional Adapter}

To complement the comparative spectral analysis of Hypothesis~2 in
\Cref{sec:qual_eva}, we employ a data-free procedure
(“Principal Component Showdown”) summarized in
Algorithm~\ref{alg:evaluation}.  This method constructs a diagnostic testbench directly from the model's weights by extracting the dominant principal directions of the
backbone’s attention projections.  
We then evaluate the conditional adapter’s response to these canonical content directions and compute the singular value spectrum of the resulting signal.

\begin{algorithm}[ht!]
\caption{Data-Free Spectral Rank Analysis (\textit{``Principal Component Showdown''})}
\label{alg:evaluation}
\DontPrintSemicolon
\SetKwInOut{Input}{Input}
\SetKwInOut{Output}{Output}
\SetKwFunction{SVD}{SVD}
\SetKwFunction{Attention}{Attention}
\SetKwFunction{TopN}{TopN}
\SetKwFunction{EffectiveRank}{EffectiveRank}

\Input{
    $W'_{q}, W'_{k}, W'_{v}$: final adapted backbone projection weights\;
    $\text{Adapter}_{\text{cond}}$: trained conditional adapter module\;
    $c_{\text{strong}}$: a strong condition value (e.g., $c = 1.0$)\;
    $N$: number of principal components to test (e.g., $N = 64$)\;
}
\Output{
    $S_{\text{text}}, S_{\text{cond}}$: normalized singular value spectra\;
    $\mathcal{R}_{\text{text}}, \mathcal{R}_{\text{cond}}$: effective ranks of
    backbone and conditional signals\;
}
\BlankLine

\tcc{1. Extract principal content primitives from the backbone}
$U_q, \_, \_ \leftarrow \SVD((W'_{q})^T)$\;
$U_k, \_, \_ \leftarrow \SVD((W'_{k})^T)$\;
$U_v, \_, \_ \leftarrow \SVD((W'_{v})^T)$\;

$q_{\text{test}} \leftarrow \TopN(U_q, N)$\;
$k_{\text{text\_test}} \leftarrow \TopN(U_k, N)$\;
$v_{\text{text\_test}} \leftarrow \TopN(U_v, N)$\;
\BlankLine

\tcc{2. Simulate backbone response to principal content directions}
$y_{\text{text\_principal}} \leftarrow
    \Attention(q_{\text{test}},
              k_{\text{text\_test}},
              v_{\text{text\_test}})$\;
\BlankLine

\tcc{3. Generate and simulate conditional response}
$K_{\text{cond}}, V_{\text{cond}} \leftarrow
    \text{Adapter}_{\text{cond}}(c_{\text{strong}})$\;
$y_{\text{cond\_principal}} \leftarrow
    \Attention(q_{\text{test}},
              K_{\text{cond}},
              V_{\text{cond}})$\;
\BlankLine

\tcc{4. Compute singular value spectra}
$\_, S_{\text{text}}, \_ \leftarrow
    \SVD(\text{flatten}(y_{\text{text\_principal}}))$\;
$\_, S_{\text{cond}}, \_ \leftarrow
    \SVD(\text{flatten}(y_{\text{cond\_principal}}))$\;
Normalize $S_{\text{text}}, S_{\text{cond}}$ by their leading singular values\;
\BlankLine

\tcc{5. Estimate effective ranks}
$\mathcal{R}_{\text{text}} \leftarrow \EffectiveRank(S_{\text{text}})$\;
$\mathcal{R}_{\text{cond}} \leftarrow \EffectiveRank(S_{\text{cond}})$\;
\BlankLine

\Return{$S_{\text{text}}, S_{\text{cond}},
        \mathcal{R}_{\text{text}}, \mathcal{R}_{\text{cond}}$}\;
\end{algorithm}

\subsection{Pyramid Sampling Strategy}

Algorithm~\ref{alg:pyramid} formalizes the pyramid generation strategy introduced in \Cref{sec:dataset_details}.
The key idea is to partition the control range $[-1,1]$ into $L$ layers with a decreasing number of scalar values per layer, then assign each layer a fixed number of independently randomized scenes.
This ensures uniform coverage of the full control range while keeping the total dataset size tractable.

\begin{algorithm}[h]
\caption{Pyramid Sampling Strategy for Dataset Construction}
\label{alg:pyramid}
\DontPrintSemicolon
\SetKwInOut{Input}{Input}
\SetKwInOut{Output}{Output}
\SetKwFunction{JitteredSample}{JitteredSample}
\SetKwFunction{GenerateScene}{GenerateScene}
\SetKwFunction{RenderScene}{RenderScene}

\Input{
    $L$: number of pyramid layers (e.g., $L=5$)\;
    $\mathbf{n} = [9, 7, 5, 3, 1]$: number of scalars per layer\;
    $S$: number of scenes per layer (e.g., $S=6$)\;
}
\Output{
    $\mathcal{D}$: training dataset of (scene, scalar, rendered clip) triples\;
}
\BlankLine
$\mathcal{D} \leftarrow \emptyset$\;
\BlankLine
\tcc{1. Iterate over pyramid layers}
\For{$l \leftarrow 1$ \KwTo $L$}{
    \BlankLine
    \tcc{2. Jittered scalar sampling: divide $[-1,1]$ into $n_l$ equal bins, sample one value per bin}
    $\mathcal{C}_l \leftarrow \JitteredSample([-1, 1],\; n_l)$\;
    \BlankLine
    \tcc{3. Generate $S$ randomized scenes for this layer}
    \For{$s \leftarrow 1$ \KwTo $S$}{
        $\text{scene}_s \leftarrow \GenerateScene()$\;
        \BlankLine
        \tcc{4. Render each scene at all scalars assigned to this layer}
        \For{$c \in \mathcal{C}_l$}{
            $\text{clip} \leftarrow \RenderScene(\text{scene}_s,\; c)$\;
            $\mathcal{D} \leftarrow \mathcal{D} \cup \{(\text{scene}_s,\; c,\; \text{clip})\}$\;
        }
    }
}
\BlankLine
\Return{$\mathcal{D}$}\;
\end{algorithm}

\section{Training Details}
All models are trained on NVIDIA A100\,80GB (A100E) GPUs using Wan2.1-T2V-14B as the backbone with bfloat16 precision. Unless noted otherwise, we follow the same configuration across all experiments. Each run trains for up to 1000 epochs with a micro-batch size of~1 per GPU, gradient accumulation of~4, and a single pipeline stage. This corresponds to a global batch size of~4 and yields roughly 18{,}000 training steps for our 30-shot synthetic dataset. Under this setup, each pyramid training stage completes in approximately one day on two A100E.

We use four condition-adapter tokens, an adapter embedding dimension of~256, and LoRA rank~32 for both the condition adapters and the backbone LoRA, with the adapter gate fixed at~0.5 throughout training. 
A learning rate of $2{\times}10^{-5}$ with a 100-step warmup is applied. Optimization is performed using the 
AdamW optimizer~\cite{loshchilov2017decoupled} with betas $(0.9,0.99)$, weight decay~0.01, and $\varepsilon=10^{-8}$. 
All LoRA and adapter parameters are trained in bfloat16 with activation checkpointing enabled for memory efficiency.


\section{Experiment Details}
\label{sec:exp_details}

In this section, we provide additional details for the experimental comparisons introduced in~\Cref{sec:experiments}.

\subsection{Group 1: Data Complexity (Syn.\ vs.\ Real)}
\label{sec:group1_details}

\begin{figure}[h]
    \centering
    \includegraphics[width=\linewidth]{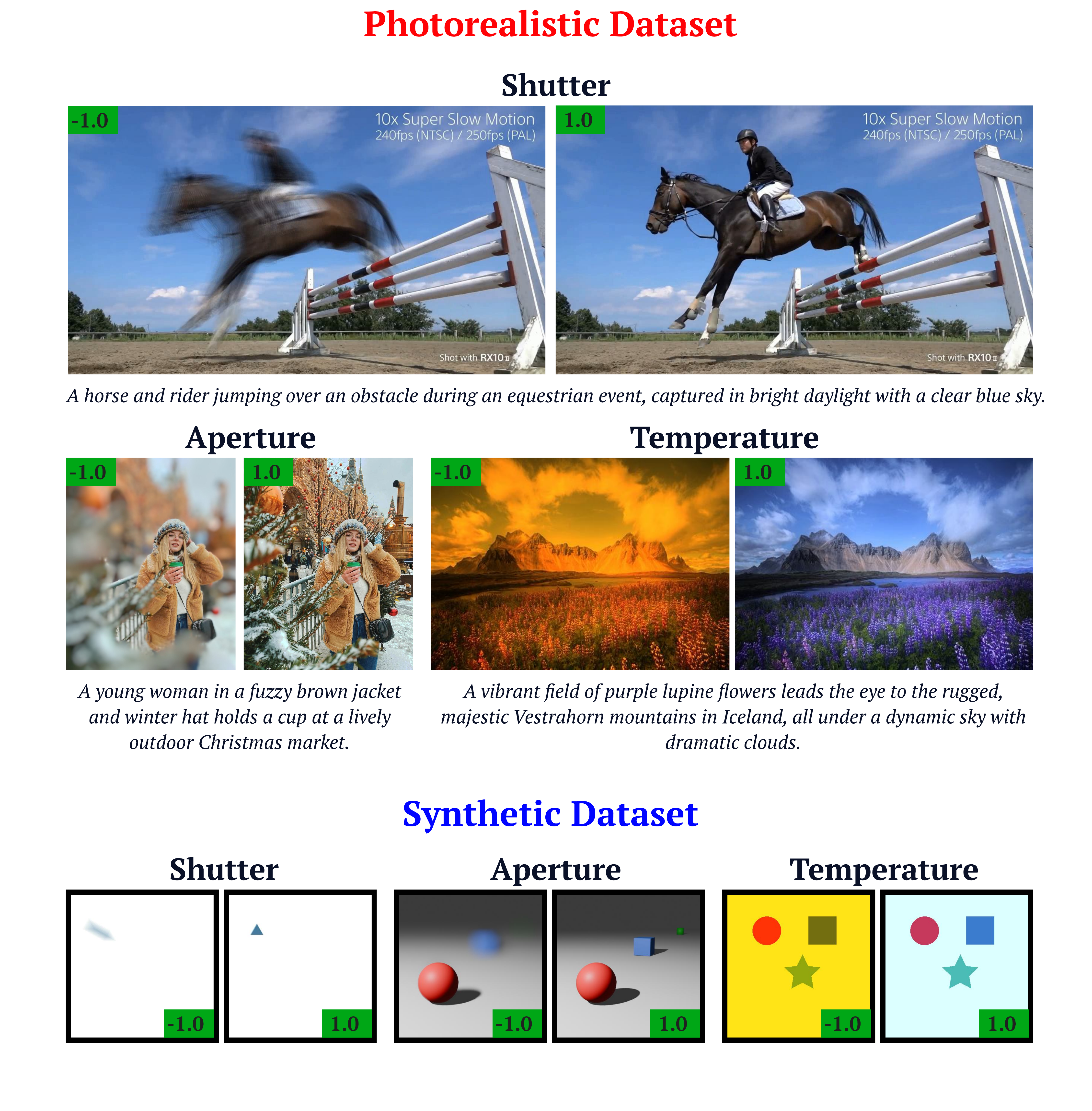}
    \caption{
Comparison of the photorealistic and synthetic one-shot datasets used in
Group~1. For each physical effect (shutter speed, aperture, temperature), we show the two extreme control values ($c=-1$ and $c=1$). Although only the endpoints are visualized, each dataset contains seven aligned conditions in total within the normalized range $c \in [-1,1]$.
    }
    \label{fig:real-syn-dataset}
\end{figure}

\noindent\textbf{Dataset Construction.} For the Group~1 comparison, we adopt a controlled one-shot setting in which each physical effect (shutter speed, aperture, temperature) is represented by a single scene containing seven scalar conditions that are aligned across the synthetic and photorealistic datasets (see~\Cref{fig:real-syn-dataset}).
All conditions are normalized to the shared range $c \in [-1,1]$, ensuring that both datasets express identical control values despite differences in rendering styles.

\vspace{1mm}
\noindent\textbf{Evaluation Prompt Details.}
Following the prompt construction strategy used in VBench---which we also rely on for the FEP metric---we generate a diverse set of 96 motion-centric T2V prompts using an LLM (Gemini~2.5). The prompts span eight categories (animals, architecture, food, humans, lifestyle, plants, scenery, and vehicles), with 12 prompts per category. We use this set to evaluate SVP under a broad range of motion patterns and scene types.

\vspace{1mm}
\noindent\textbf{Quantitative Results Recap.}
As shown in~\Cref{fig:quan-compare}, models fine-tuned on our simple synthetic dataset exhibit substantially smaller drift from the pretrained backbone when monitored using the FEP metrics (\Cref{sec:fep}). In contrast, training on the photorealistic dataset leads to noticeably larger and faster deviation throughout training. This difference in drift is reflected in the final SVP results (\Cref{sec:svp}): the model trained on synthetic data maintains near-baseline semantic fidelity, while the model trained on photorealistic data shows clear backbone degradation. These findings highlight that, for learning physical controls, low-complexity synthetic data provides a more stable and
robust supervision signal.

\begin{figure*}[t]
    \centering
    \includegraphics[width=\linewidth]{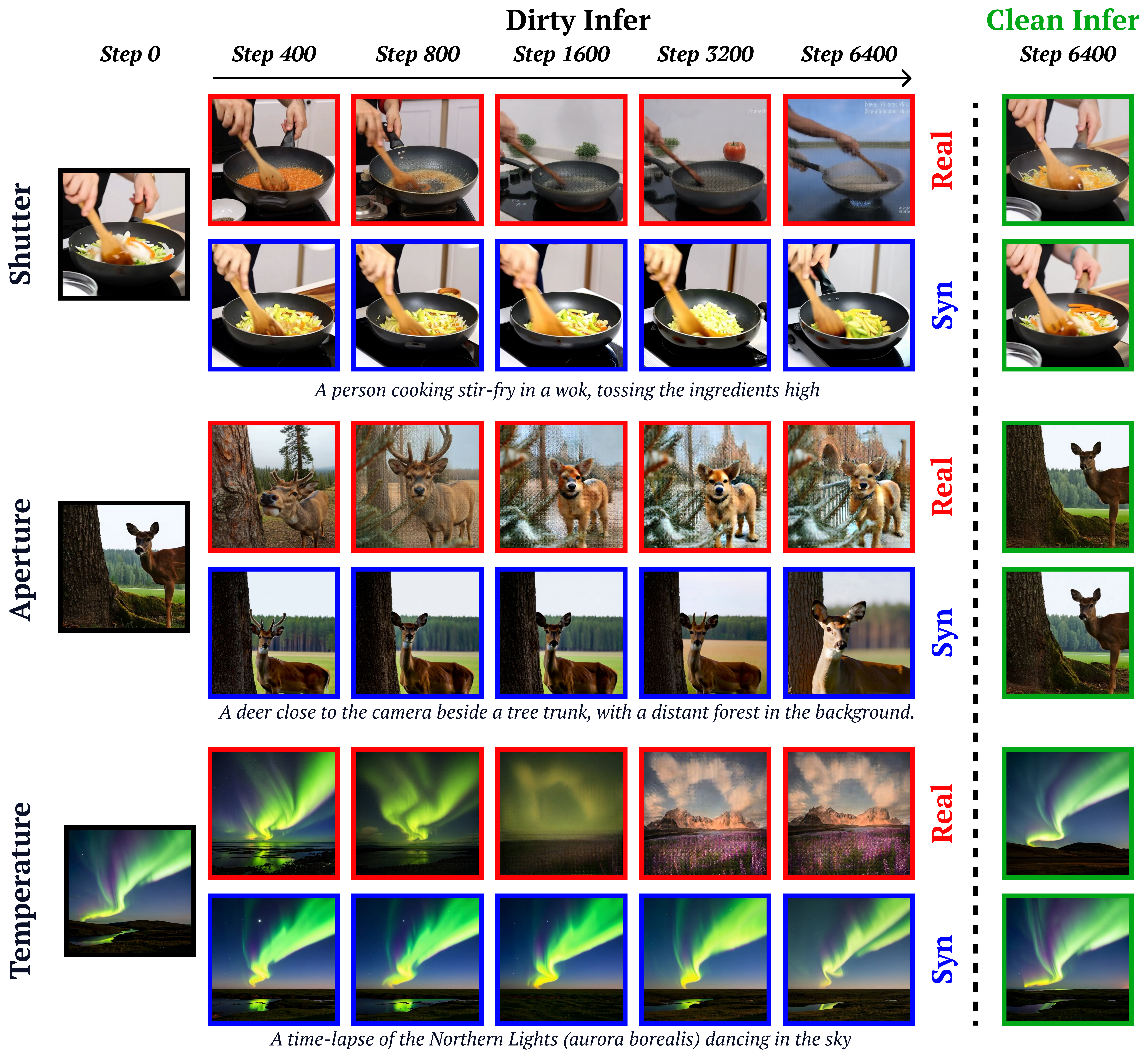}
    \caption{
       Visual comparison of backbone corruption during training. The
        synthetic-trained model remains stable and continues to follow the input prompt, whereas the model trained on photorealistic data rapidly drifts toward copying its training scene (see \cref{fig:real-syn-dataset}). This manifests as semantic drift, texture collapse, and global color shifts that intensify over training.
    }
    \label{fig:visual-corrupt}
\end{figure*}

\vspace{1mm}
\noindent\textbf{Visualization Evaluation.} Beyond the quantitative trends, the backbone degradation becomes most apparent when inspecting the model’s inference outputs during training (see~\Cref{fig:visual-corrupt}). 
When trained on our synthetic dataset, the model continues to follow the input prompt and preserves the overall appearance of the generated scene. In contrast, the model trained on the photorealistic dataset begins to visually drift toward the training images themselves: details,
textures, and global color tones increasingly resemble the underlying
photorealistic scene rather than the prompted content. 
This ``copying'' behavior appears early in training and progressively intensifies, manifesting as semantic drift, texture collapse, and color hallucination. These visual failures provide an intuitive counterpart to the backbone corruption measured by our FEP and SVP metrics.

\subsection{Group 2: Inference Strategy (Decoupled vs. Full-LoRA)}
\label{sec:group2_details}

While Group~1 highlights the stability advantage of synthetic data during
training, here we analyze how inference strategy further affects the final output quality.
Specifically, we compare \textbf{Full-LoRA Inference}, which retains all backbone LoRA weights at test time, against our proposed \textbf{Decoupled Inference}, which prunes the shallow DiT blocks to remove residual content drift before decoding.

\vspace{1mm}
\noindent\textbf{Quantitative Results Recap.}
Similar to the quantitative trends observed in Group~1, the metrics in
Group~2 also reveal a clear advantage for Decoupled Inference. As summarized in \Cref{tab:svp_results} and the corresponding FEP/SVP curves and charts in \Cref{fig:quan-compare}, Decoupled Inference consistently yields lower drift and higher semantic fidelity compared to Full-LoRA Inference across all physical controls. Even when the model is trained on the synthetic dataset—where backbone corruption is already minimal—Decoupled Inference further reduces the residual deviations and restores the model’s output closer to the original WAN~2.1 behavior. These quantitative results motivate a closer look at the visual differences between the two inference modes.

\vspace{1mm}
\noindent\textbf{Visual Evidence.} The visual comparisons further reinforce the quantitative trends. As shown in the last column of \Cref{fig:visual-corrupt}, applying Decoupled Inference to a corrupted checkpoint trained on photorealistic data dramatically restores prompt fidelity and global appearance: the model no longer copies textures or colors from the training scene, and semantic drift is substantially reduced.
This demonstrates that much of the degradation observed during training is not permanently baked into the weights but instead arises from residual content leakage that Decoupled Inference effectively removes.

\begin{figure}[t]
    \centering
    \includegraphics[width=\linewidth]{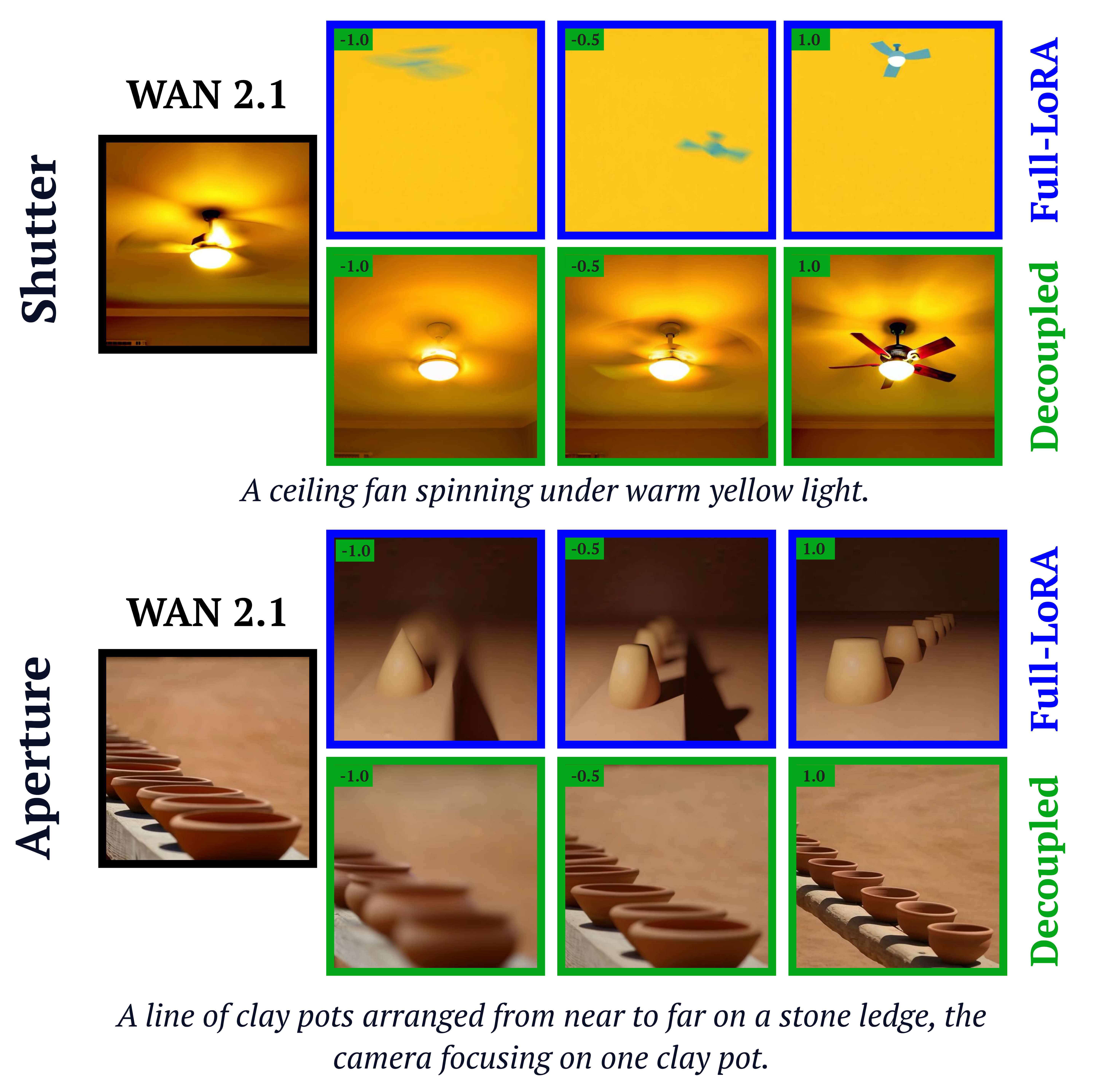}
    \caption{
    Effect of Decoupled vs.\ Full-LoRA Inference on a model trained with the synthetic pyramid dataset.
    Although synthetic training induces only mild backbone drift, Full-LoRA Inference still injects subtle synthetic-style biases (e.g., flattened shading and softened textures) into the outputs.
    Applying Decoupled Inference removes these artifacts and restores the visual style of the original WAN~2.1 model across different control values.
    }
    \label{fig:clean-syn}
\end{figure}

A similar pattern appears even when the model is trained on the synthetic
dataset. Although backbone corruption is minimal in this setting, and the model is capable of yielding high-quality output as shown in \Cref{fig:backbone_infer}, Full-LoRA Inference still injects subtle synthetic-style bias—such as flattened textures or blurred shading—into the final output. As shown in \Cref{fig:clean-syn}, Decoupled Inference suppresses these artifacts and produces videos that closely match the visual style and sharpness of the original WAN~2.1 backbone across different control values.

\begin{figure}[t]
    \centering
    \includegraphics[width=\linewidth]{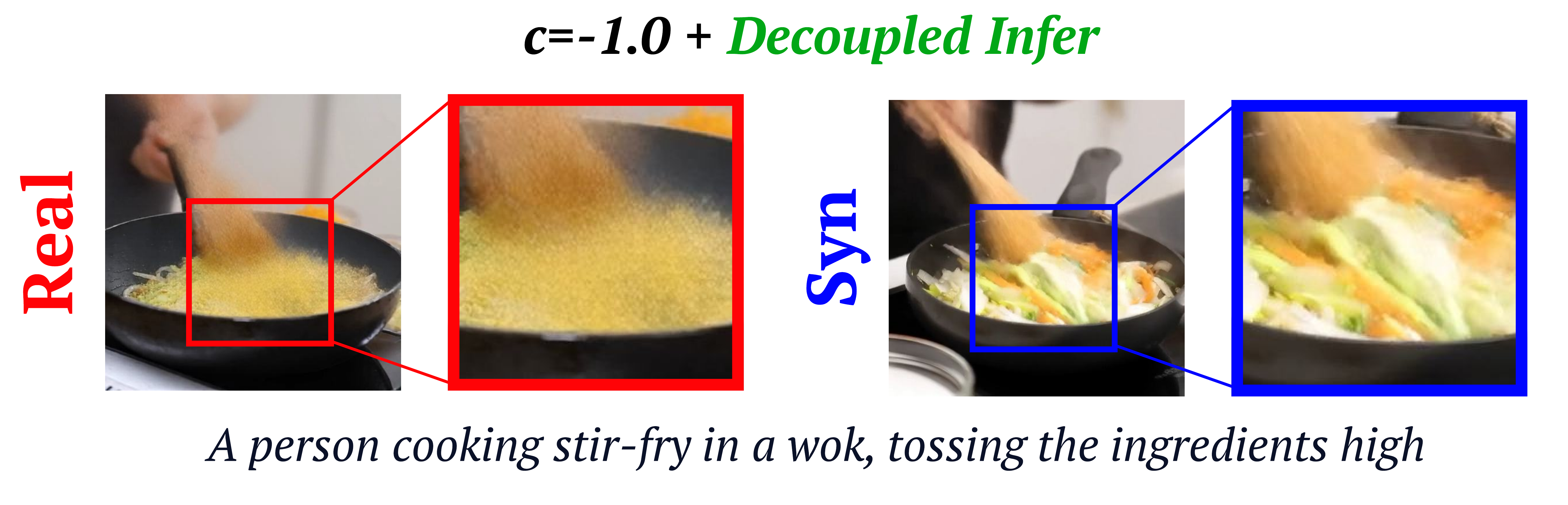}
    \caption{
       Residual artifacts under Decoupled Inference at extreme control values ($c=-1.0$, shutter speed). Even after pruning the shallow LoRA blocks, the model trained on real data (left) exhibits visible noise and structural degradation in high-motion regions (red inset), owing to the complex, high-entropy content of the real training sequence. The model trained on synthetic data (right) produces a clean motion-blur effect under the same condition (blue inset), consistent with its simpler, more disentangled supervision signal.
    }
    \label{fig:real-fails}
\end{figure}

\vspace{1mm}
\noindent\textbf{Limitations of Real Data.} Although Decoupled Inference substantially reduces the corruption introduced by real-data training, it does not fully eliminate it—especially under extreme control values.
As shown in \Cref{fig:real-fails}, the model trained on the
real shutter speed dataset still exhibits noise and structural artifacts when the control approaches the boundaries (e.g., $c=-1.0$), even after applying Decoupled Inference. This occurs because the real sequence contains complex high-entropy motion: articulated limbs, nonlinear trajectories, and strong background parallax. 
In contrast, the synthetic shutter speed dataset is constructed from simple rigid motion with clean parametric trajectories, providing a much more disentangled supervision signal. 

These observations reinforce our \emph{Less is More} thesis: low-complexity synthetic data is inherently easier for the model to factorize and scales reliably across the full control range, whereas real data introduces irreducible entanglement that Decoupled Inference can mitigate but not fully correct. 
Thus, synthetic data remains the preferred choice for learning stable,
well-behaved physical controls.

\section{Quantitative Controllability Analysis}
\label{sec:monotonicity}

To quantitatively validate the precision and monotonicity of our learned controls, we perform a post-hoc analysis on existing 5-point sweeps $c \in \{-1, -0.5, 0, 0.5, 1\}$ across $N$ diverse prompts per control (no additional training required).
For each frame, we compute a lightweight image-level proxy capturing the intended physical effect: \textbf{shutter} uses Laplacian-variance sharpness (inversely proportional to blur strength); \textbf{aperture} uses background-region sharpness (inversely proportional to defocus); \textbf{temperature} uses a global warm--cool color-cast statistic.
We then measure the Spearman rank correlation $\rho$ between the control scalar $c$ and the proxy across the 5-point sweep for each prompt.

Results are reported in \Cref{tab:monotonicity}.
The median $|\rho|$ is 1.00 across all three controls, with 100\% correct direction in every case, confirming that the learned controls produce consistent, monotonic responses to the conditioning scalar.
The slightly lower mean $|\rho|$ for aperture (0.912) reflects occasional non-perfect adjacent steps attributable to proxy sensitivity at fine-grained defocus levels or scene-content variation; the corresponding sweep videos in \Cref{sec:gallery} confirm that the visual effect remains well-behaved throughout.

\begin{table}[h]
  \centering
  \caption{
    \textbf{Quantitative Monotonicity Analysis.}
    Spearman rank correlation $|\rho|$ between the control scalar $c$ and a per-effect image proxy, computed over 5-point sweeps ($c \in \{-1,-0.5,0,0.5,1\}$). $N$ is the number of prompts evaluated per control. Direction accuracy is 100\% for all controls.
  }
  \label{tab:monotonicity}
  \begin{tabular*}{\columnwidth}{@{\extracolsep{\fill}}lccc@{}}
    \toprule
    \textbf{Control} & $\mathbf{N}$ & \textbf{Median $|\rho|$} & \textbf{Mean $|\rho|$} \\
    \midrule
    Shutter Speed & 15 & 1.000 & 0.987 \\
    Aperture      & 17 & 1.000 & 0.912 \\
    Temperature   & 14 & 1.000 & 1.000 \\
    \bottomrule
  \end{tabular*}
\end{table}

\section{Out-of-Range Inference}
\label{sec:ood_inference}

\begin{figure*}[h]
  \centering
  \includegraphics[width=\linewidth]{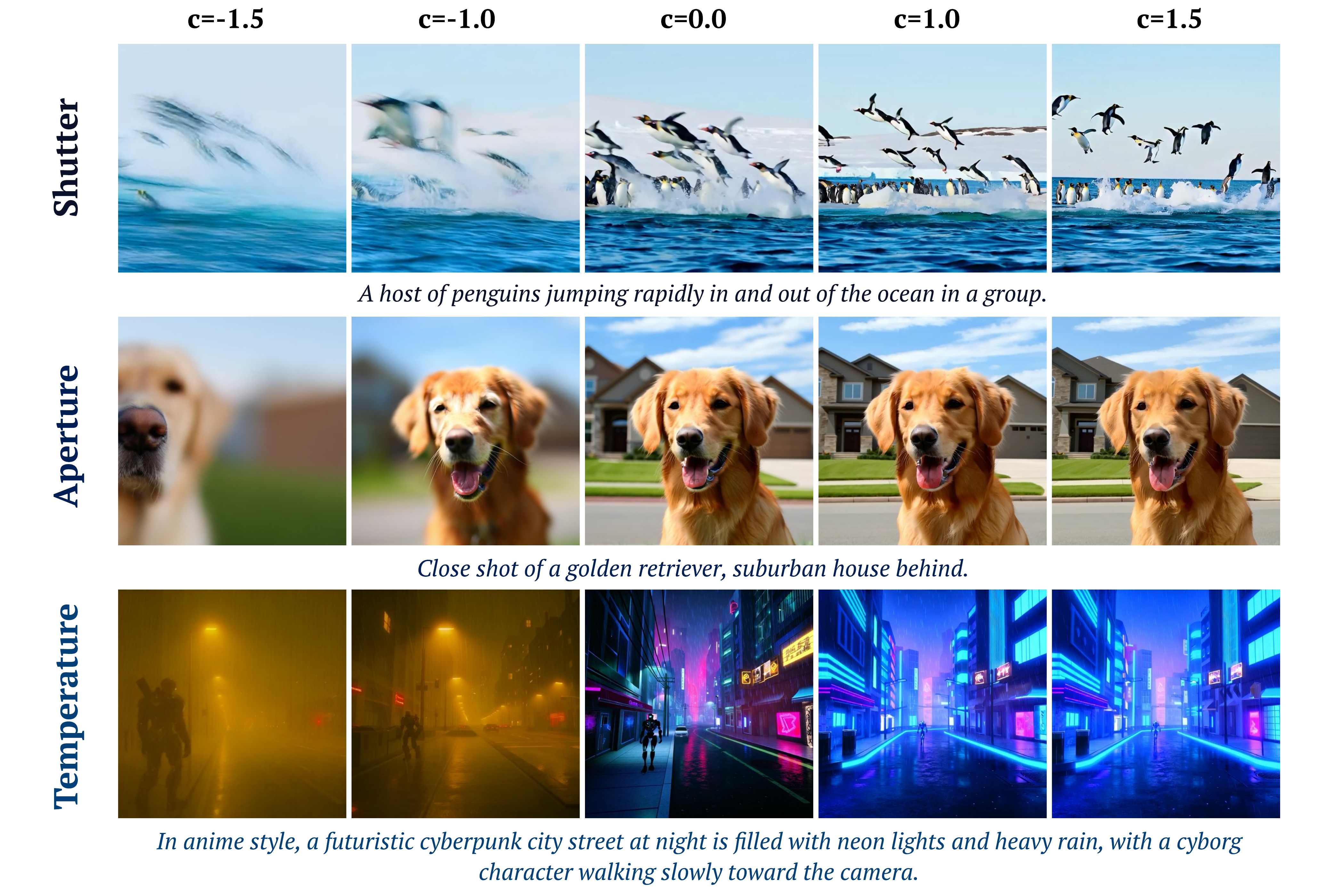}
  \caption{
    \textbf{Out-of-range inference.}
    Qualitative sweep over $c \in \{-1.5, -1.0, 0.0, 1.0, 1.5\}$ for shutter speed, aperture, and color temperature.
    Values $c = \pm 1.5$ lie outside the training range $[-1, 1]$.
    The model extrapolates plausibly in all three cases, with the physical effect strengthening monotonically and prompt fidelity preserved throughout.
  }
  \label{fig:ood_inference}
\end{figure*}

A natural question is how the model behaves when the conditioning scalar $c$ is pushed beyond the training range $[-1, 1]$.
\Cref{fig:ood_inference} shows qualitative results for $c \in \{-1.5, -1.0, 0.0, 1.0, 1.5\}$ across all three controls, where $c = \pm 1.5$ falls outside the training distribution.

The model exhibits \textbf{plausible extrapolation}: the intended physical effect continues to strengthen smoothly beyond the training boundary without degrading prompt fidelity or visual coherence.
For shutter speed, $c = -1.5$ produces stronger motion blur while scene content remains coherent; for aperture, $c = -1.5$ yields more extreme bokeh with the subject still well-rendered; for color temperature, the warm--cool gradient extends naturally at both extremes.
This behavior is consistent with successful low-dimensional disentanglement: the learned conditioning operates on a compact, continuous manifold that generalizes modestly beyond its training support.
Probing the limits of this extrapolation---and characterizing where prompt guidance eventually degrades at more extreme values---is an interesting direction for future work.

\section{Analysis Details}
\subsection{Hypothesis 1: The Spectrum of Backbone Adaptation}
\label{sec:heatmap}

\begin{figure*}[t]
  \centering
  \begin{subfigure}{0.45\linewidth}
    \includegraphics[width=\linewidth]{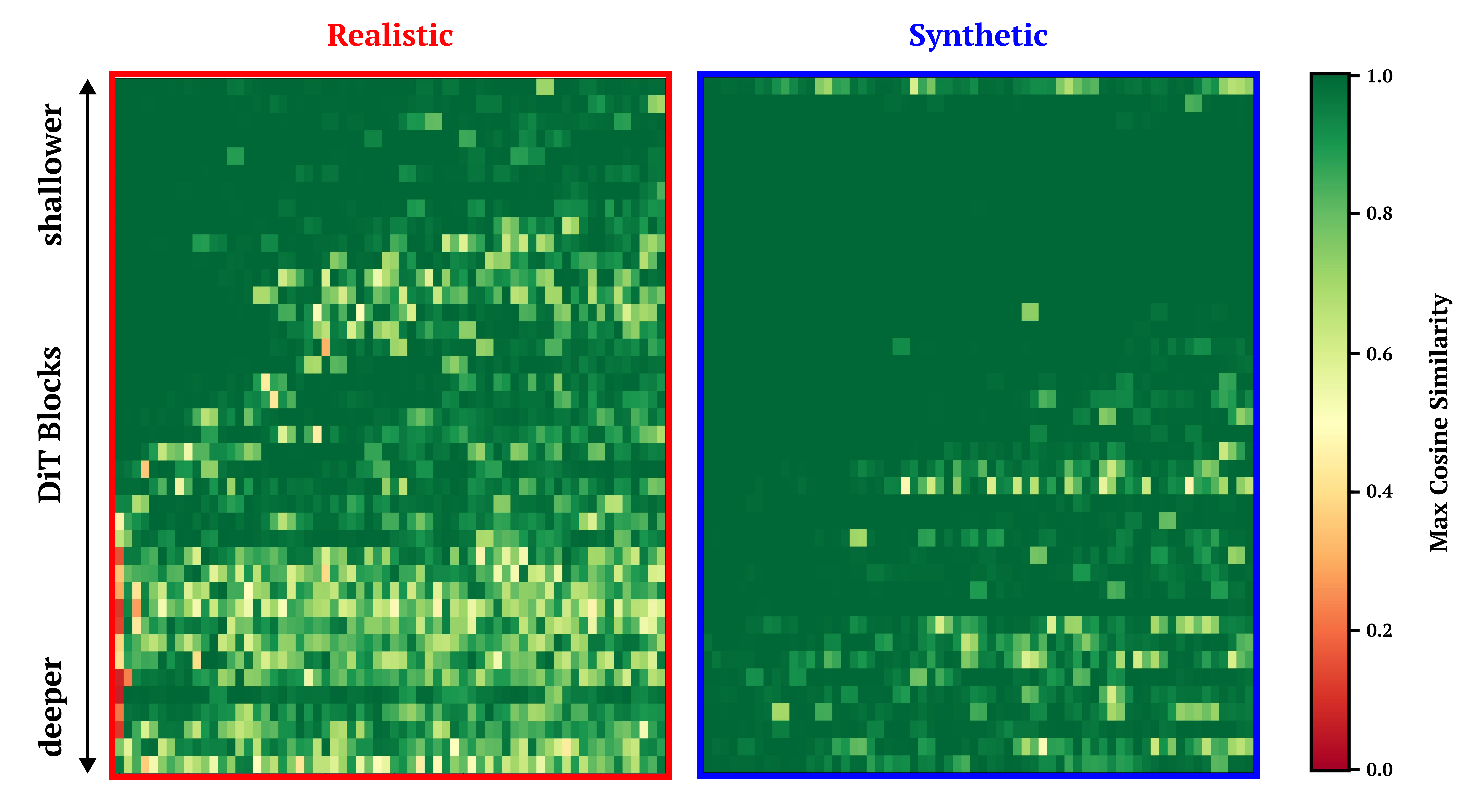}
    \caption{Similarity Heatmap Comparison}
    \label{fig:heatmap_comparison}
  \end{subfigure}
  \hfill
  \begin{subfigure}{0.45\linewidth}
    \includegraphics[width= \linewidth]{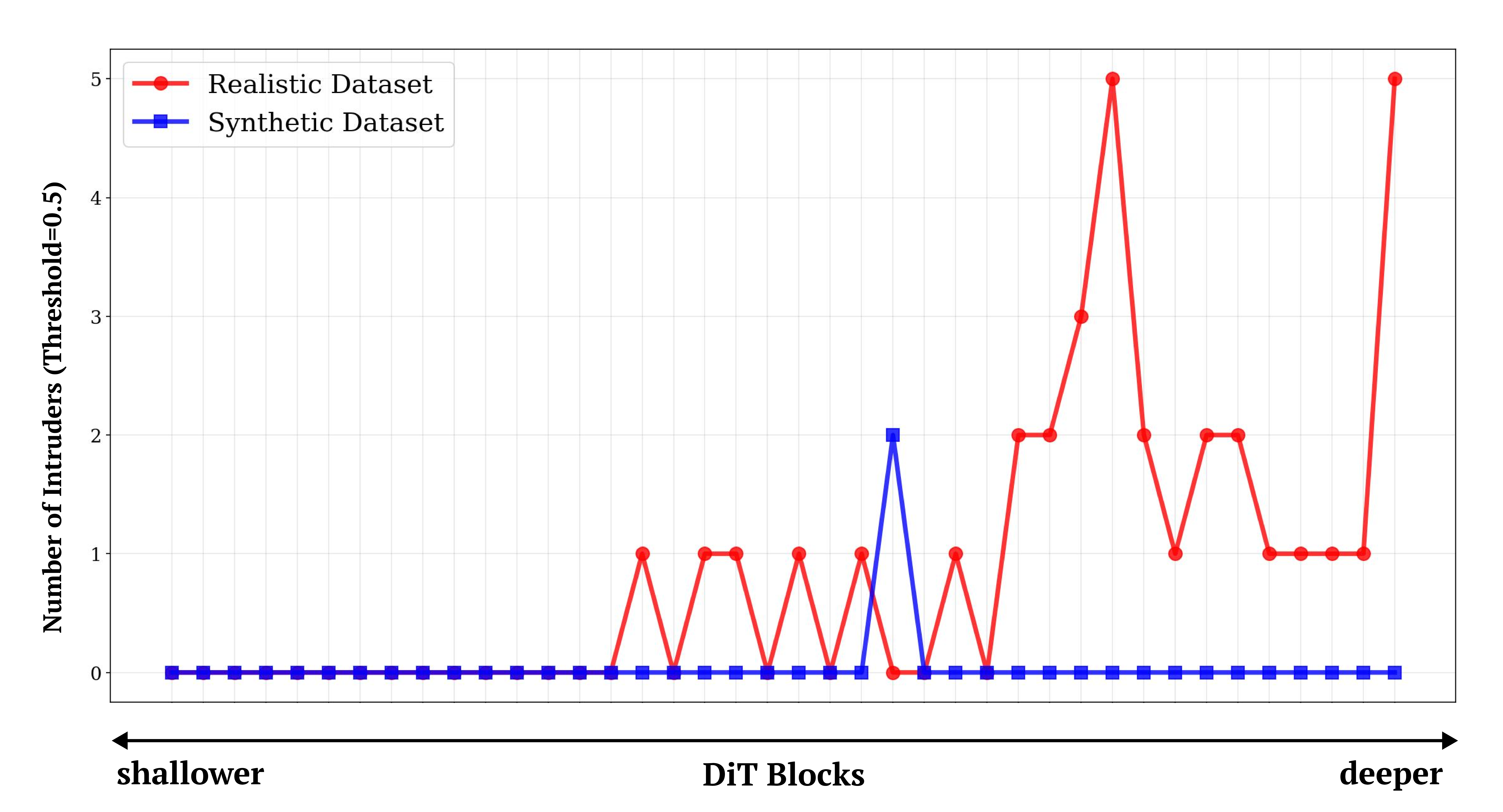}
    \caption{Intruder Dimension Comparison}
    \label{fig:intruder_comparison}
  \end{subfigure}
  \caption{Backbone content drift across depth for the shutter speed condition. Analysis performed on the value projection \(W_v\).
(a) Max cosine similarity between content directions and FPS-conditioned features.
(b) Number of content intruders (cosine $>0.5$).  }
\label{fig:hyp1_analysis}
\end{figure*}

Expanding on Hypothesis~1 in \Cref{sec:qual_eva}, we analyze how the complexity of fine-tuning data influences the backbone’s spectral drift under conditional adaptation. Unless otherwise noted, all measurements are performed on the value projection \(W_v\), though we observe qualitatively similar trends in \(W_q, W_k,\) and \(W_o\). Figure~\ref{fig:hyp1_analysis} reports the results for the shutter speed condition using both real and synthetic training data as discussed in \cref{sec:exp_details}.
Other physical controls exhibit the same overall behavior:
simple synthetic data preserves the pretrained spectral
structure, whereas complex real data drives substantial high-rank
intrusions into the backbone.

\paragraph{Clean Adaptation (Single Synthetic Scene).}
As shown in Fig.~\ref{fig:heatmap_comparison}, synthetic shutter speed conditioning results in minimal drift throughout the network. The similarity heatmap remains uniformly low (green), indicating that the adapter modifies \(W_v\) through small non-destructive adjustments. This corresponds to a clean and stable form of adaptation: the model adjusts to the new condition while largely preserving the pretrained spectral structure. However, the specialization remains narrow,
as the synthetic data contains only one simple scene.

\paragraph{Catastrophic Forgetting (Single Complex Scene).}
In contrast, Fig.~\ref{fig:intruder_comparison} reveals that real shutter
conditioning induces substantial spectral drift in \(W_v\), particularly in shallow and mid-depth blocks. A large number of high-similarity intruders (cosine similarity $>0.5$) appear as early as blocks 0--12 and accumulate into dense clusters in deeper layers. This is the hallmark of catastrophic forgetting: the model learns strong, scene-specific features that overwrite general-purpose representations. These intruder directions propagate into the conditioned path during inference, producing the ``ghosting’’ artifacts seen in the qualitative
comparisons. Similar behavior appears under other real-data conditions.

Overall, this data-free analysis corroborates our quantitative findings and highlights that deeper blocks inherently provide cleaner, more stable channels for condition modulation.

\begin{figure*}[t]
    \centering
    \includegraphics[width=0.98\linewidth]{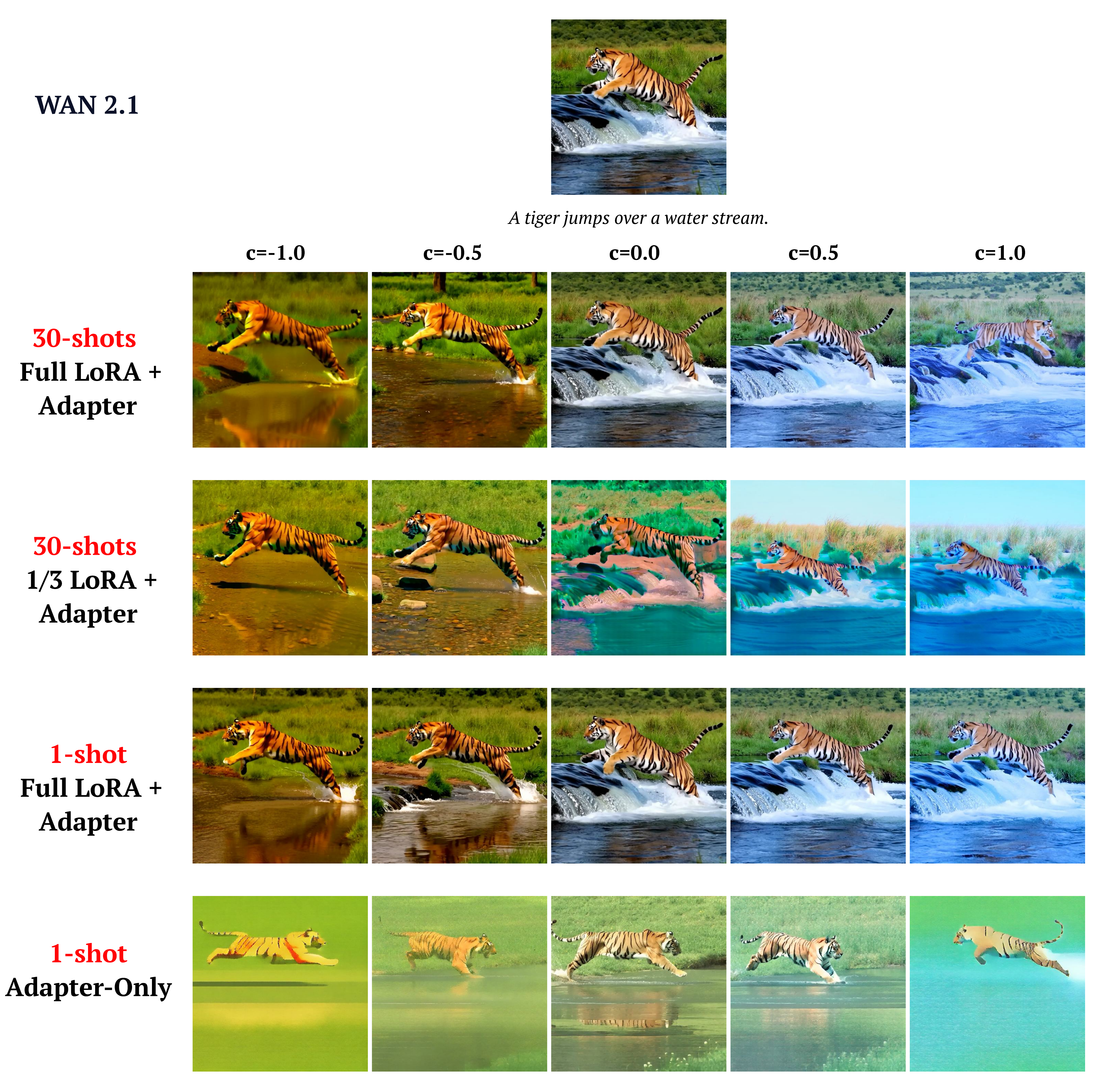}
    \caption{
        \textbf{Qualitative comparison of training configurations on the temperature control (tiger scene).}
        The first row shows the pre-trained Wan2.1 backbone output without conditioning, serving as the visual reference for generative fidelity.
        The second row (our proposed Joint Training on the full pyramid synthetic dataset) demonstrates non-destructive adaptation: at $c=0$ the output is visually indistinguishable from the backbone reference, while varying $c$ produces clean, isolated temperature changes without scene content drift.
        The third row (1/3 LoRA + Adapter) captures the conditioning effect but introduces an unnatural warm bias at positive control values, indicating that without a full backbone LoRA to absorb the synthetic domain shift, a distributional bias contaminates the conditioning pathway.
        The fourth row (one-shot Joint Training) preserves backbone fidelity comparably to the second row, but exhibits a slightly weaker conditioning effect.
        The fifth row (Adapter-Only training) exhibits the ``Bulldozer Effect'': scene content is dominated by training-data features regardless of the condition value.
    }
    \label{fig:hyp2_vis}
\end{figure*}

\subsection{Hypothesis 2: Full Backbone LoRA Is Essential for Context-Disentangled Conditioning}
\label{sec:low_rank_vis}

In Hypothesis~2 from \Cref{sec:qual_eva}, our spectral analysis demonstrated that the jointly trained model produces a low-rank, effect-specific conditional signal, whereas an adapter-only model produces a high-rank signal that conflates the physical effect with training-scene content.
Here we extend this comparison to four training configurations, each evaluated on the same temperature-conditioned tiger scene, to provide a comprehensive visual validation of why Joint Training with a full backbone LoRA constitutes the correct architectural choice.
The objective is not to assess the photorealism of the temperature effect in isolation, but to determine whether each configuration can express the physical control cleanly without corrupting scene content or degrading the backbone's generative priors.

\paragraph{Joint Training, Full Pyramid Dataset (Proposed).}
As shown in the second row of \Cref{fig:hyp2_vis}, our proposed method satisfies both objectives.
At $c=0$, the output is visually indistinguishable from the backbone reference (first row), providing direct visual evidence that joint training is non-destructive and preserves the original generative priors.
As $c$ varies, the temperature shift is applied in a content-stable manner, consistent with the compact, low-rank conditional signal identified in \Cref{fig:rank_comparison}.

\paragraph{Partial Backbone Coverage: 1/3 LoRA + Adapter.}
The third row restricts the backbone LoRA to the deepest third of the DiT blocks---mirroring the block coverage used at inference time, now applied during training.
While the temperature effect is successfully induced, an unnatural warm color bias emerges at positive control values.
This reveals that the full backbone LoRA is necessary during training to absorb the synthetic domain shift across all network depths; without it, the unabsorbed distributional bias propagates into the conditioning pathway and manifests as a persistent, condition-correlated color offset.

\paragraph{Data Efficiency: One-Shot Joint Training.}
The fourth row applies the full joint training configuration to a single training scene.
Backbone fidelity is well preserved, confirming that the Joint Training architecture remains non-destructive even under minimal data.
The conditioning effect is slightly weaker compared to training on more scenes, suggesting that a more diverse training set tends to produce more robust physical conditioning; a systematic study of data scaling is left to future work.

\paragraph{Adapter-Only Training and the Bulldozer Effect.}
The fifth row trains only the conditional adapter without a backbone LoRA.
In the absence of a dedicated pathway for absorbing domain shift, the adapter is forced to encode both the physical effect and the training-scene context within its conditional pathway.
As described in \Cref{sec:qual_eva}, this produces the ``Bulldozer Effect'': the high-rank, high-magnitude conditional signal suppresses the text-based content signal, causing the model's output to be dominated by training-scene features irrespective of the conditioning scalar.

Taken together, these four configurations confirm a consistent finding: Joint Training with a full backbone LoRA is the only configuration that simultaneously achieves accurate physical conditioning and non-destructive backbone adaptation, while increasing training scene diversity further strengthens the robustness of the learned control.

\section{Example Gallery}
\label{sec:gallery}

\begin{figure*}[t]
\centering
\includegraphics[width= 0.95\textwidth]{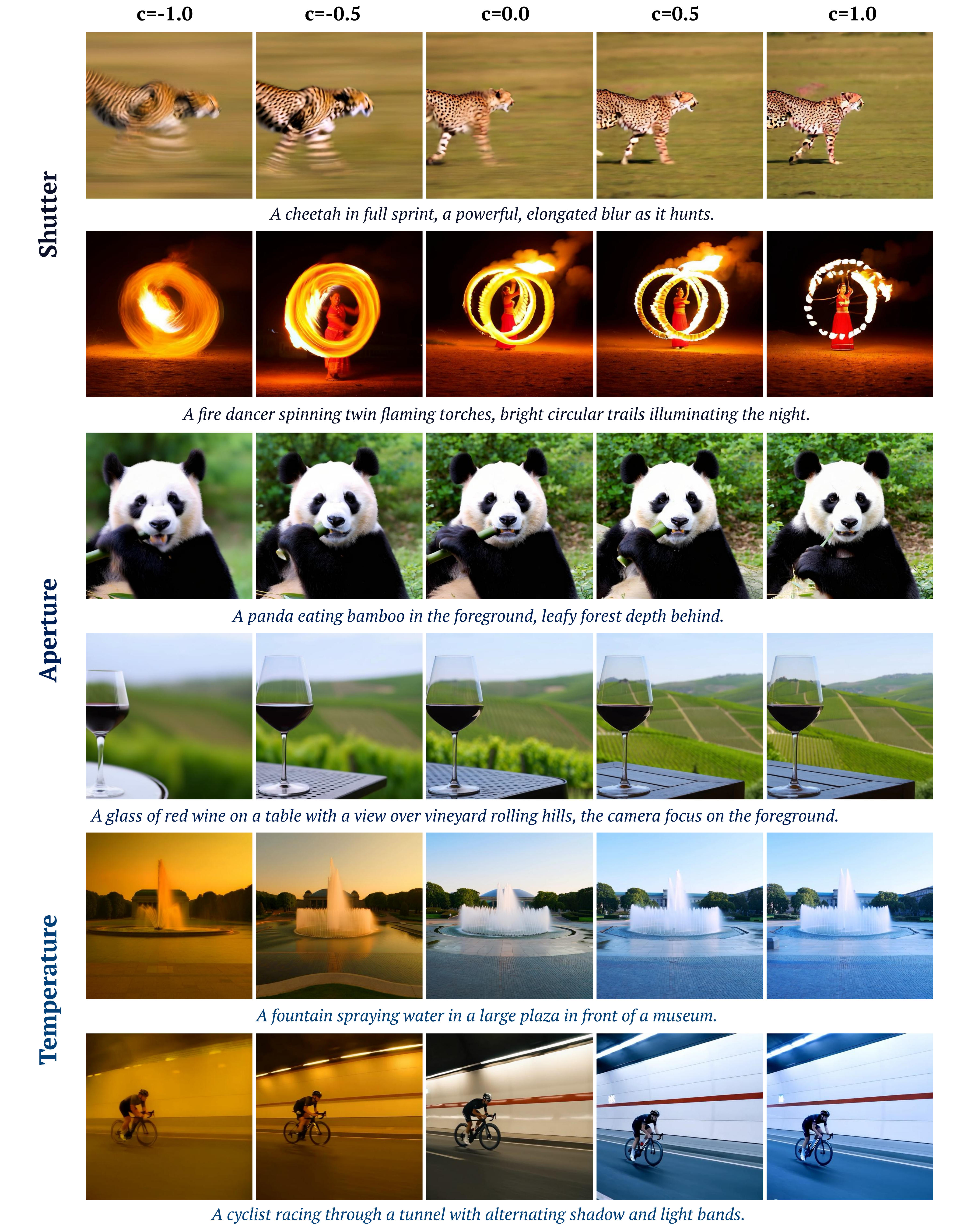}
\caption{\textbf{Qualitative results of our controllable generation.} Our model demonstrates precise and continuous control over shutter speed (Rows 1--2, motion blur), aperture (Rows 3--4, bokeh), and color temperature (Rows 5--6) by varying the conditional input $c$ from -1.0 to 1.0 across diverse, high-fidelity video prompts.}
\label{fig:gallery}
\end{figure*}

Figure~\ref{fig:gallery} presents a diverse set of examples generated with our shutter-, aperture-, and temperature-controlled T2V framework. Each physical control is parameterized within a unified range $c \in [-1,1]$, enabling smooth and interpretable adjustments to motion blur, depth of field, and color tone. Across a wide variety of scenes and prompts, the outputs change in a consistent and approximately monotonic manner as $c$ varies, demonstrating the reliability and generality of our learned control space.

\begin{figure*}[t]
\centering
\includegraphics[width=\textwidth]{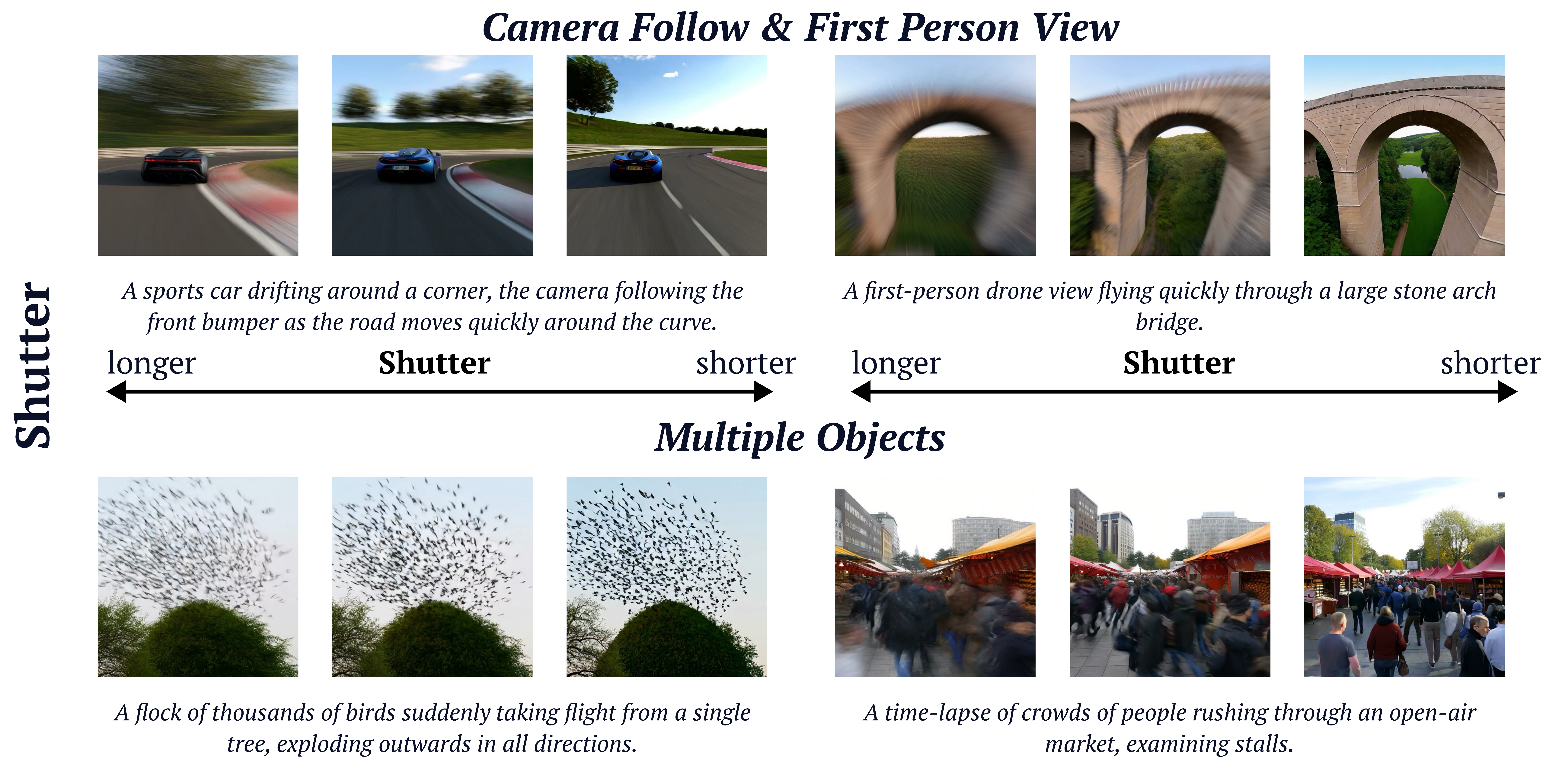}
\caption{\textbf{Generalization of shutter control to scenes with complex motion.} The model responds reliably to the shutter scalar in settings involving moving cameras (e.g., camera-follow and first-person views) and scenes with multiple independently moving objects.}
\label{fig:shutter-gen}
\end{figure*}

\begin{figure*}[t]
\centering
\includegraphics[width=\textwidth]{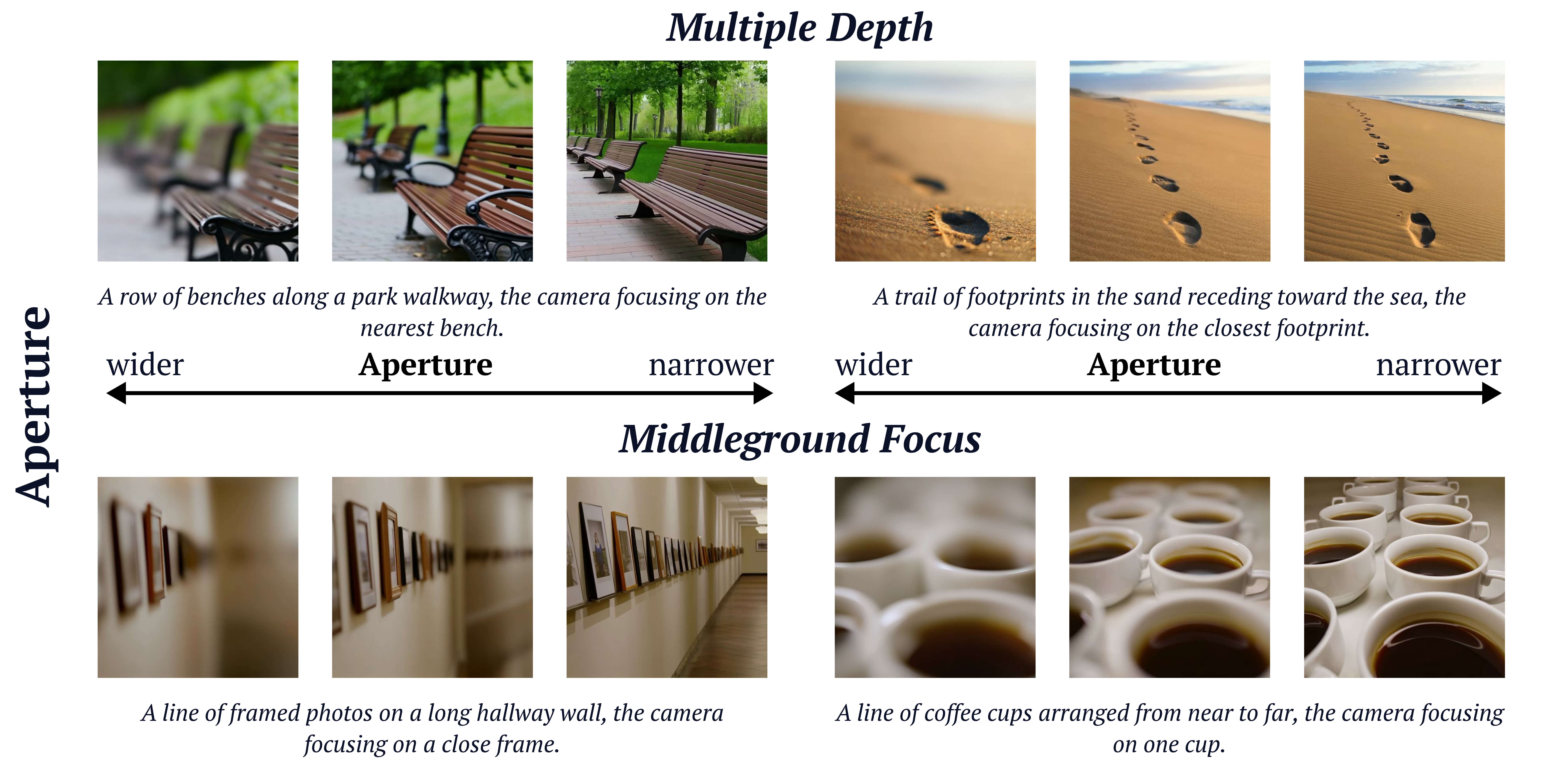}
\caption{\textbf{Generalization of aperture control across diverse depth layouts and focal targets.} The model handles scenes with multiple depth layers and varied spatial arrangements, and can focus on locations beyond the foreground (e.g., mid- or
background planes).}
\label{fig:aperture-gen}
\end{figure*}

\begin{figure*}[t]
\centering
\includegraphics[width=\textwidth]{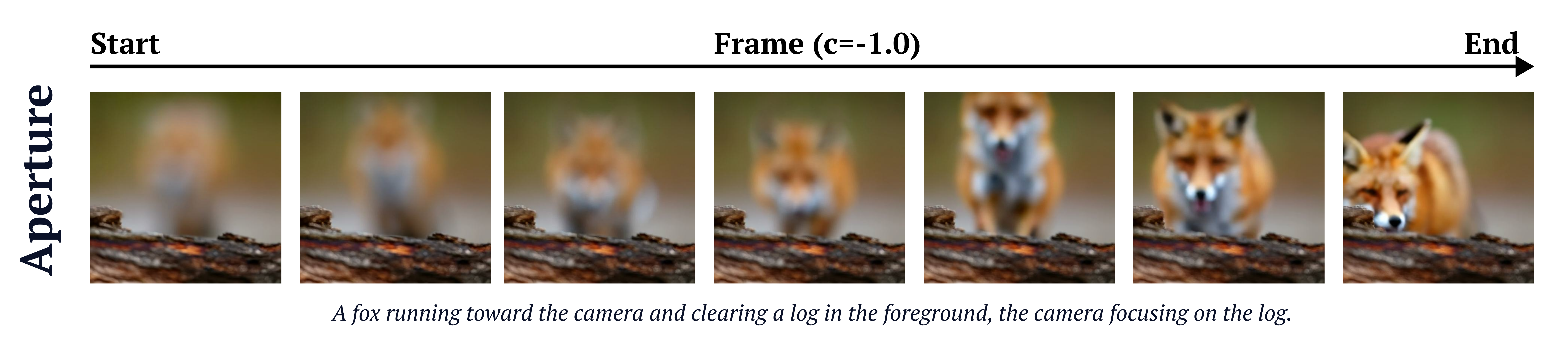}
\caption{Despite being trained only on images, the model renders smooth bokeh variation as depth changes, enabled by the backbone’s strong prior.}
\label{fig:aperture-move}
\end{figure*}

\begin{figure*}[t]
\centering
\includegraphics[width=\textwidth]{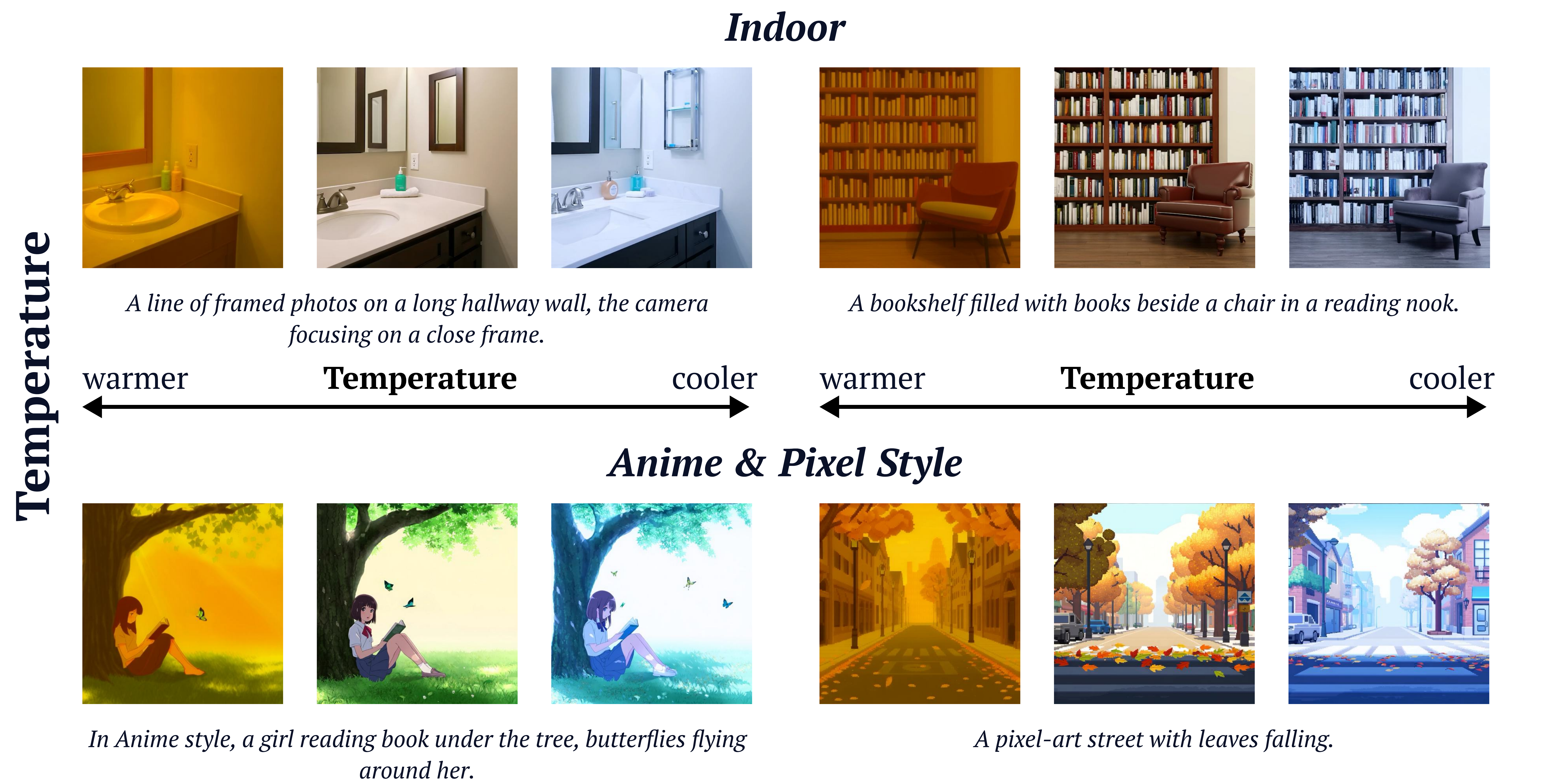}
\caption{\textbf{Generalization of temperature control across a wide range of scene types.} The model produces stable cooler-to-warmer transitions in indoor environments as well as highly stylized domains such as anime and pixel art.}
\label{fig:temp-gen}
\end{figure*}

To further assess this generalization, Figures~\ref{fig:shutter-gen},~\ref{fig:aperture-gen}, and~\ref{fig:temp-gen} evaluate each control dimension in more challenging scenarios. Shutter control extends to moving-camera settings and scenes with multiple independently moving objects, while maintaining predictable blur variation. Aperture control remains stable across diverse depth layouts and can shift focus to locations beyond the foreground, producing smooth bokeh shifts as depth varies (see \Cref{fig:aperture-move}).  Temperature control remains consistent across indoor environments and highly stylized domains such as anime
and pixel art. Together, these results show that our learned controls
generalize far beyond the simple one-shot synthetic scenes used for training.

{
    \small
    \bibliographystyle{ieeenat_fullname}
    \bibliography{main}
}

\end{document}